\documentclass[sn-mathphys]{sn-jnl}% Math and Physical Sciences Reference Style
\usepackage[caption=false,font=normalsize,labelfont=sf,textfont=sf]{subfig}
\jyear{2022}%

\theoremstyle{thmstyleone}%
\newtheorem{theorem}{Theorem}[section]% meant for sectionwise numbers
\theoremstyle{thmstyletwo}%

\theoremstyle{thmstylethree}%
\newtheorem{assumption}[theorem]{Assumption}

\newcommand{\eg}{\emph{e.g.},\ }
\newcommand{\ie}{\emph{i.e.},\ }

\newcommand{\bs}{\mathbf{s}}
\newcommand{\ba}{\mathbf{a}}

\begin{document}

\title[STRAPPER]
{STRAPPER: Preference-based Reinforcement Learning via Self-training Augmentation and Peer Regularization}


\author[1,2]{\fnm{Yachen} \sur{Kang}}\email{kangyachen@westlake.edu.cn}

\author[2]{\fnm{Li} \sur{He}}\email{heli@westlake.edu.cn}

\author[2]{\fnm{Jinxin} \sur{Liu}}\email{liujinxin@westlake.edu.cn}

\author[2]{\fnm{Zifeng} \sur{Zhuang}}\email{zhuangzifeng@westlake.edu.cn}

\author*[2]{\pfx{Prof. Dr.} \fnm{Donglin} \sur{Wang}}\email{wangdonglin@westlake.edu.cn}

\affil[1]{\orgdiv{College of Computer Science and Technology}, \orgname{Zhejiang University}, \orgaddress{\city{Hangzhou}, \postcode{310024}, \state{Zhejiang}, \country{China}}}

\affil*[2]{\orgdiv{Machine Intelligence Lab (MiLAB) of the School of Engineering}, \orgname{Westlake University}, \orgaddress{\city{Hangzhou}, \postcode{310024}, \state{Zhejiang}, \country{China}}}


\abstract{
Preference-based reinforcement learning (PbRL) promises to learn a complex reward function with binary human preference.
However, such human-in-the-loop formulation requires considerable human effort to assign preference labels to segment pairs, hindering its large-scale applications.
Recent approache has tried to reuse unlabeled segments, which implicitly elucidates the distribution of segments and thereby alleviates the human effort.
And consistency regularization is further considered to improve the performance of semi-supervised learning.
However, we notice that, unlike general classification tasks, in PbRL there exits a unique phenomenon that we defined as \emph{similarity trap} in this paper.
Intuitively, human can have diametrically opposite preferredness for similar segment pairs, but such similarity may trap consistency regularization fail in PbRL.
Due to the existence of \emph{similarity trap}, such consistency regularization improperly enhances the consistency possiblity of the model's predictions between segment pairs, and thus reduces the confidence in reward learning, since the augmented distribution does not match with the original one in PbRL.
% \emph{Similarity trap} cause consistency regularization to improperly improve the possibly of model's prediction consistency between disjoint segment pairs and reduces the confidence in reward learning, since the augmented distribution does not match with the original one in PbRL.
To overcome such issue, we present a self-training method along with our proposed peer regularization, which penalizes the reward model memorizing uninformative labels and acquires confident predictions.
Empirically, we demonstrate that our approach is capable of learning well a variety of locomotion and robotic manipulation behaviors using different semi-supervised alternatives and peer regularization.
}




\maketitle

\section{Introduction}\label{sec1}

Deep reinforcement learning (RL) utilizes a flexible framework for learning task-oriented behaviors \cite{NateKohl2004PolicyGR,JensKober2008PolicySF,JensKober2013ReinforcementLI,DavidSilver2017MasteringTG,DmitryKalashnikov2018QTOptSD,OriolVinyals2019GrandmasterLI}, where the design of the reward function is commonly necessary and crucial.
However, many tasks are often complex or hard to be specified so that the reward design is very difficult.
Agents often learn to exploit some loopholes of a misspecified reward function, resulting in unwanted behaviors (misaligned with the original intention).
In addition, requirements such as operational safety and compliance with social norms are difficult to effectively state and meet through reward engineering \cite{DarioAmodei2016ConcretePI,RohinShah2019PreferencesII,AlexanderMattTurner2020AvoidingSE}.

% Imitation learning \cite{TianhaoZhang2018DeepIL}, alleviating the excessive reliance of human designer on sophisticated reward engineering, promises to directly imitate or infer the potential reward functions from suitable demonstrations.
% Imitation learning \cite{SylvainCalinon2009LearningCM,PeterPastor2011OnlineMA,BarisAkgun2012TrajectoriesAK,TianhaoZhang2018DeepIL}, alleviating the excessive reliance of human designer on sophisticated reward engineering, promises to directly imitate or infer the potential reward functions from suitable demonstrations.
% However, these formulations still require non-trivial human effort to make sure demonstrations remain expert \cite{SylvainCalinon2009LearningCM,PeterPastor2011OnlineMA,BarisAkgun2012TrajectoriesAK,TianhaoZhang2018DeepIL}.
Preference-based RL (PbRL) \cite{PaulFChristiano2017DeepRL,ErdemByk2018BatchAP,DorsaSadigh2017ActivePL,ErdemByk2020ActivePG,KiminLee2021BPrefBP}, as an alternative, provides a paradigm to elicit a reward function from human preference between two trajectory segments.
PbRL permits learning a flexible reward function with only binary preferences, which requires less non-trivial human effort.
However, in standard RL, learning from such sparse human prior is a notorious challenge.
Compared with imitation learning, PbRL does not require demonstrators to be expert.
Unfortunately, the main benefit of PbRL, relying on neither hand-crafted rewards nor expert demonstrators, is also what makes it costly or impractical.
Due to the ingredient of human-in-the-loop, the straightforward PbRL inevitably requires a large amount of human feedbacks, which is costly and hinders its large-scale applications.
To enable more scalable and practical human-in-the-loop learning, prior works tried to make this process more efficient in terms of human feedback, by designing various strategies to select informative queries \cite{PaulFChristiano2017DeepRL, shin2021offline, KiminLee2021PEBBLEFI, liang2021reward, BorjaIbarz2018RewardLF}.

Compared to human preferences, segments can be accessed more easily, and a large amount of unlabeled segments can help implicitly elucidate the distribution of segments \cite{chapelle2009semi}.
So in this paper, we focus on \emph{how to reuse the unlabeled segments} to help learning the reward function.
It is common to apply semi-supervised learning (SSL) on the dataset that can be augmented by adding unlabeled segments.
% Orthogonal to prior works aiming to select informative queries, in this paper, we particularly focus on \emph{how to reuse the unlabeled segments} to help learning the reward function, so as to reduce the amount of human preferences needed.
% Reusing the unlabeled segments can help to implicitly elucidate the distribution of segments \cite{chapelle2009semi}.
% To instantiate this, it is a common practice to apply the semi-supervised learning (SSL) methods to mitigate the requirement for labeled segments by providing a means of augmenting unlabeled segments.
Some attempts have been made in SURF \cite{park2022surf} to apply simple semi-supervised learning method (pseudo-labeling) to PbRL, and have been proven to be effective for improving the efficiency of preference samples.

\begin{figure*}[t] %htbp
	\centering
	\includegraphics[width=0.88\textwidth]{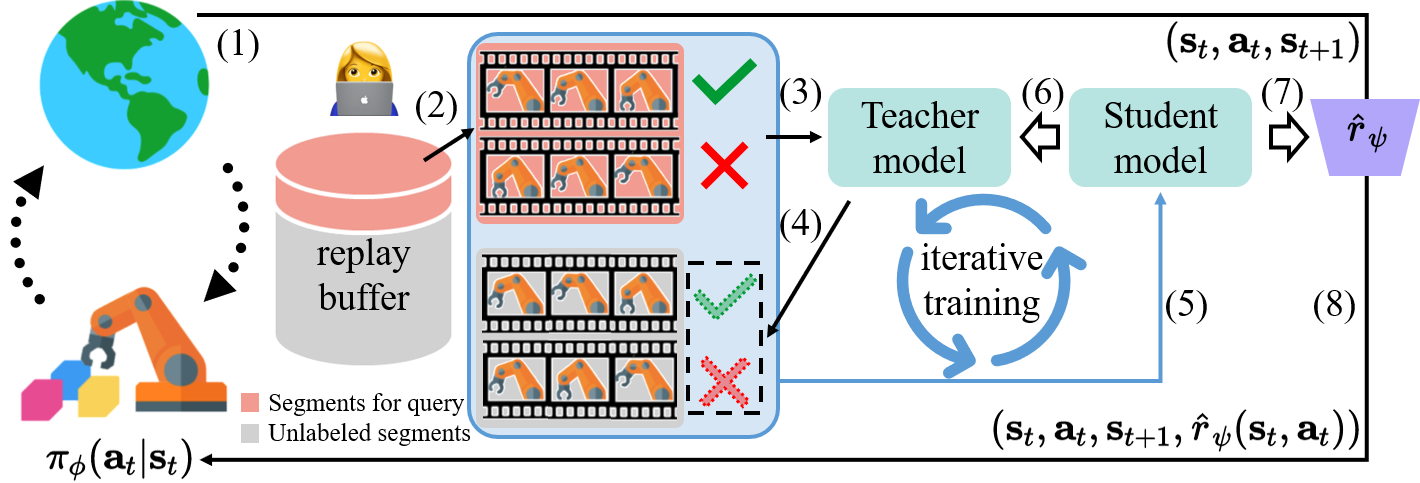}
	% \vspace{-10pt}
	\caption{
		Illustration of STRAPPER.
		(1) Generate new transtions.
		(2) Sample segment pairs to query preferences from human.
		(3) Train teacher model using labeled data.
		(4) Pseudo-label unlabeled data using teacher model.
		(5) Train student model with consistency and peer regularization using mixed data.
		(6) Use the student model as the new teacher model.
		(7) Use the reward function trained in the student model to reward the transitions.
		(8) Update the policy.
	}
	\label{illustration}
	% \vspace{-10pt}
\end{figure*}

A widely-used SSL method is self-training with consistency regularization (CR) \cite{laine2016temporal, sajjadi2016regularization}.
Specifically, self-training has three main steps in PbRL:
1) train a teacher model on labeled segments (labeled dataset),
2) use the teacher to generate pseudo-labels on unlabeled segments (unlabeled dataset), and
3) train a student model on the combination of labeled and pseudo-labeled segments (mixed dataset),
and then self-training treats the student as a teacher to relabel the unlabeled data and iterates the three steps above.
Consistency regularization encourages the model to have identical output for different augmented data, by constraining model prediction invariant to input noise \cite{xie2019unsupervised}.
Most SOTA SSL methods adopt consistency regularization as an additional loss component; however, so far it has not been fully discussed in PbRL.

We observe that in PbRL there is a unique phenomenon we named as \emph{similarity trap}, and such phenomenon hinders the direct use of consistency regularization.
Specifically, in PBRL, humans can derive completely opposite preference labels from only a few, but fatal, differences.
When there exits such data in unlabeled dataset, the conventional SSL method would cause model to improperly improve the consistency of predictions between disjoint data and thus reduce the confidence in reward learning.
To counteract such a negative effect, we present a new PbRL framework: \textbf{S}elf-\textbf{TR}aining \textbf{A}ugmented \textbf{P}reference-based learning via \textbf{PE}er \textbf{R}egularization, abbreviated as STRAPPER (See Figure \ref{illustration}).
We use an iterative self-training procedure to exploit the trajectory data without preference labels, resulting in a reward function.
And we add peer regularization in self-training process, discouraging the student model from outputting the same label for peer samples (two samples that independently sampled in mixed dataset) inspired by~\cite{YangLiu2020PeerLF}.
% So we add a peer regularization term to counteract, discouraging the student model from outputting the same labels for the peer samples (independently sampled two data in mixed dataset).
Intuitively, peer samples would not provide any informative signals to each other.
Thus, we hope to penalize the student model to memorize uninformative labels, encouraging more confident prediction.

% To summarize, we present a new preference-based reinforcement learning framework: \textbf{S}elf-\textbf{TR}aining \textbf{A}ugmented \textbf{P}reference-based learning via \textbf{PE}er \textbf{R}egularization, abbreviated as STRAPPER.
% Our method relies on two main, synergistic ingredients: self-training and peer regularization (for the specific steps, see Figure \ref{illustration}).
Here, we summarize the main contribution of STRAPPER as:
% % \item We propose a framework to reuse the unlabeled segments in PbRL and show the potential to reduce the human effort in PbRL using semi-supervised alternatives while keeping compatable performance.
% (1) We observe and propose a phenomenon called \emph{similarity trap} that hinders direct applying consistency regularization to PbRL.
% (2) We propose peer regularization to fix such issue and empirically verify it.
% (3) STRAPPER consistently outperforms prior PbRL baselines on complex locomotion and robotic manipulation tasks from DeepMind Control Suite and Meta-world.
% (4) We empirically verify that our method can also be useful for preference-based RL to eliminate the (potential) noise in preferences.
% % \item STRAPPER outperforms prior PbRL baselines on complex locomotion and robotic manipulation tasks from DeepMind Control Suite \cite{tassa2018deepmind} and Meta-world \cite{yu2021metaworld}.

\begin{itemize}
	\item We propose a framework to reuse the unlabeled segments in PbRL and show the potential to reduce the human effort in PbRL using semi-supervised alternatives while keeping competitive performance.
	\item We observe a crucial phenomenon in PbRL called \emph{similarity trap} that hinders direct applying consistency regularization to PbRL. To address this, we propose a novel peer regularization to fix such issue and empirically verify it.
	\item STRAPPER consistently outperforms prior PbRL baselines.
	In addition, we show that our method can generally eliminate the potential noise in preferences.
	% \item STRAPPER outperforms prior PbRL baselines on complex locomotion and robotic manipulation tasks from DeepMind Control Suite \cite{tassa2018deepmind} and Meta-world \cite{yu2021metaworld}.
\end{itemize}

\section{Related Work}
\subsection{Preference-based RL.}
PbRL is one of RL paradigms to learn from human feedback \cite{BorjaIbarz2018RewardLF}.
Several works have successfully utilized feedback from real human to train RL agents \cite{DilipArumugam2019DeepRL,PaulFChristiano2017DeepRL,BorjaIbarz2018RewardLF,WBradleyKnox2009InteractivelySA,KiminLee2021PEBBLEFI,GarrettWarnell2017DeepTI}.
\cite{PaulFChristiano2017DeepRL} scales PbRL to utilize modern deep learning techniques, and \cite{BorjaIbarz2018RewardLF} improves the efficiency of this method by introducing additional forms of feedback such as demonstrations.
Recently, \cite{KiminLee2021PEBBLEFI} proposes a feedback-efficient RL algorithm by utilizing off-policy learning and pre-training.
% To improve sample- and feedback-efficiency of human-in-the-loop RL, previous works \cite{PaulFChristiano2017DeepRL,BorjaIbarz2018RewardLF,KiminLee2021PEBBLEFI,JanLeike2018ScalableAA} mainly focus on methods such as selecting more informative queries \cite{PaulFChristiano2017DeepRL} and pre-training of RL agents \cite{BorjaIbarz2018RewardLF,KiminLee2021PEBBLEFI}.
\cite{park2022surf} uses pseudo-labeling to utilize unlabeled segments and proposes a novel augmentation called temporal cropping to augment labeled data.

% {\color{red}
% Our proposed approach
% % similarly focuses on developing a more sample- and feedback-efficient preference-based RL algorithm without adding any additional forms of supervision.
% % Instead, we enable off-policy learning as well as utilize unsupervised pre-training to substantially improve efficiency.
% % Stiennon et al. [61] and Wu et al. [71] showed that preference-based RL can be utilized to fine-tune GPT-3 [16] for hard tasks like text and book summarization, respectively.
% % We benchmark these state-of-the-art preference-based RL algorithms in this paper.
% % We further investigate effects of different exploration methods in preference-based RL algorithm.
% % We follow a common approach of exploration methods in RL: generating intrinsic rewards as exploration bonus Pathak et al. (2019).
% % Instead of only using learned reward function from human feedback as RL training objective, we alter the reward function to include a combination of the extrinsic reward (the learned rewards) and an intrinsic reward (exploration bonus).
% % In particular, we present an exploration method with intrinsic reward that measures the disagreement from learned reward models.
% }

\subsection{Semi-supervised learning.}
Self-training denotes a learning setting where the supervision on unlabeled data is given by the own prediction of the model trained with the labeled data \cite{DavidYarowsky1995UNSUPERVISEDWS,KamalNigam2000AnalyzingTE,zhu2005semi,CharlesJRosenberg2005SemiSupervisedSO}.
Pseudo-labeling refers to a specific variant, where model predictions are converted to hard labels \cite{lee2013pseudo}.
This is often used along with a confidence-based threshold that retains unlabeled examples only when the classifier is sufficiently confident.
Consistency regularization was first proposed by \cite{bachman2014learning} and later referred to as the “$\Pi$-Model” \cite{rasmus2015semi}, where the main idea is to force the output of the model remain constant with randomly augmented inputs.
After that, FixMatch \cite{sohn2020fixmatch} first combines consistency regularization and pseudo-labeling in one simple method.
On the other hand, NoisyStudent \cite{QizheXie2020SelfTrainingWN} employs a self-training scheme, a teacher model generates pseudo-labels on unlabeled data, and a larger noisy student model is then trained on both labeled and unlabeled data with consistency regularization.
Recently, \cite{zhu2021rich} proposes an analytical framework to unify consistency regularization with explicit and implicit pseudo-labels.
\section{Preliminaries}
\subsection{Reinforcement learning.}
We consider a standard RL framework where an agent interacts with an environment in discrete time \cite{RichardSutton1988ReinforcementLA}.
% We consider an agent interacting with an environment in discrete time .
Formally, at each timestep $t$, the agent receives a state $\bs_t$ from the environment and chooses an action $\ba_t$ based on its policy $\pi_\phi$.
% At each timestep $t$, the agent receives a state $s_t$ from the environment and chooses an action at based on its policy $\pi$.
% The environment returns a reward $r_t$ and the agent transitions to the next state $s_{t+1}$.
In traditional RL, the environment also returns a reward $r(\bs_t,\ba_t)$.
The return $\mathcal{R}_t = \sum^\infty_{k=0} \gamma^k r_{t+k}$ is the discounted sum of rewards from timestep $t$ with discount factor $\gamma \in (0,1]$.
RL then maximizes the expected return $\mathcal{R}_0$ with respect to $\pi_\phi$.

\subsection{Preference-based RL (PbRL).}
However, for many complex domains and tasks, it is difficult to construct a suitable reward function.
We consider the PbRL framework, where human provide preferences between two behavior segments and then the agent trains under this supervision \cite{PaulFChristiano2017DeepRL,BorjaIbarz2018RewardLF,KiminLee2021PEBBLEFI,JanLeike2018ScalableAA,KiminLee2021BPrefBP}.
In this work, we follow the classic framework to learn a reward function $\widehat r_\phi$ from preferences, where the function is trained to be consistent with human feedback \cite{AaronWilson2012ABA,PaulFChristiano2017DeepRL}.
In this framework, a segment $\sigma$ is a sequence of states and actions $\{\bs_h,\ba_h,\dots,\bs_{h+H},\ba_{h+H}\}$.
Transitions $(\bs_t,\ba_t,\bs_{t+1})$ are stored in a replay buffer $\mathcal{B}$, and we sample segments $\sigma$ from $\mathcal{B}$.
% Formally, a segment $\sigma$ is a sequence of observations and actions $\{(s_1,a_1),\dots,(s_H,a_H)\}$.
Then, we elicit preferences $y$ for segments $\sigma^i$ and $\sigma^j$.
% More specifically, given a pair of segments $(\sigma^i,\sigma^j)$, human annotator indicates which segment is preferred, \ie $y = (0,1)$ or $(1,0)$, or that the two segments are equally preferred, \ie $y = (0.5,0.5)$, or that two segments are incomparable, \ie discarding the query.
More specifically, $y$ indicates which segment is preferred, i.e., $y \in \{0,1,0.5\}$, where $0$ indicates $\sigma^i \succ \sigma^j$ (the event that segment $\sigma^i$ is preferable to $\sigma^j$), $1$ indicates $\sigma^j \succ \sigma^i$ ($\sigma^j$ is preferable to $\sigma^i$), and $0.5$ implies an equally preferable case.
The judgment is recorded in a dataset $D$ as a triple $(\sigma^i,\sigma^j,y)$.

By following the Bradley-Terry model \cite{Bradley1952RankAO}, we have a preference predictor as follows:
% \small
\begin{equation}\label{classifier}
    P_{\psi}[\sigma^{i} \succ \sigma^{j}]
    =\frac{\exp \sum_{t} \widehat{r}_{\psi}(\mathbf{s}_{t}^{i}, \mathbf{a}_{t}^{i})}
    % {\exp \sum_{t} \widehat{r}_{\psi}(\mathbf{s}_{t}^{i}, \mathbf{a}_{t}^{i})
    % +\exp \sum_{t} \widehat{r}_{\psi}(\mathbf{s}_{t}^{j}, \mathbf{a}_{t}^{j})}
    {\sum_{m \in\{i,j\}} \exp \sum_{t} \widehat{r}_{\psi}(\mathbf{s}_{t}^{m}, \mathbf{a}_{t}^{m})} ,
\end{equation}
% \endsmall
where $\widehat r_\psi$ is the reward function.
% where $\sigma^i \succ \sigma^j$ denotes the event that segment $\sigma^i$ is preferable to segment $\sigma^j$.
Intuitively, we assume that the probability of preferring one segment is exponentially proportional to the sum of an underlying reward function over the segment.
While $\widehat r_\psi$ is not a binary classifier, learning $\widehat r_\psi$ amounts to binary classification with labels $y$ provided by an annotator.
Concretely, the reward function, modeled as a neural network with parameters $\psi$, is updated by minimizing the following loss:
\begin{equation}\label{loss}
    \begin{split}
        \mathcal{L}^{\text {Reward}}
        =-\underset{(\sigma^{i}, \sigma^{j}, y) \sim D}{\mathbb{E}}
        \Big[y(0) \log P_{\psi}[\sigma^{i} \succ \sigma^{j}]
            +y(1) \log P_{\psi}[\sigma^{j} \succ \sigma^{i}]\Big],
    \end{split}
\end{equation}
% \begin{equation}\label{loss}
%     \begin{split}
%         \mathcal{L}^{\text {Reward}}
%         =-\underset{(\sigma^{i}, \sigma^{j}, y) \sim D}{\mathbb{E}}
%         &\Big[y(0) \log P_{\psi}[\sigma^{i} \succ \sigma^{j}]\\
%             &+y(1) \log P_{\psi}[\sigma^{j} \succ \sigma^{i}]\Big].
%     \end{split}
% \end{equation}
where $y(0)$ and $y(1)$ represent whether the preference label is 0 or 1, respectively.
Finally, we use the learned reward function $\widehat r_\psi$ to train the policy.

\subsection{Semi-supervised Learning in PbRL.}
Here, we model PbRL into the framework of the semi-supervised learning.
The labeled dataset $D_L$ is first formed with segment pairs $(\sigma^{i}, \sigma^{j})$ sampled from trajectory buffer $\mathcal{B}$ and the preferences $y$ queried from human, \eg $D_L:=\{(\sigma^{i}, \sigma^{j},y)\}$.
% The corresponding clean distribution is also denoted by $\mathcal{D}$, where $(\sigma^{i}, \sigma^{j}, y)$ are drawn from.
% $P_{\psi}$ in Eq.\ref{classifier} is the classifier model we would like to optimize.
We use $D_U$ to denote the dataset formed with segment pairs $(\sigma^{i}, \sigma^{j}, \cdot)$ without human labels.

\section{\emph{Similarity Trap} in PbRL}\label{problem}
% \subsection{Reusing Unlabeled Segments in PbRL}
% The goal of preference-based RL is to train an agent to perform human-desired behaviors using as few queries as possible, since it requires expensive human efforts.
% Previous works use pre-training to increase the efficiency of the human initial feedback.
% Collecting a breadth of experiences enables the human to provide more meaningful feedback, as compared to feedback on data collected in an indeliberate manner.
% % However, we claim that unlabeled segments are much easier to obtain than segments labeled by preference, since it require less human efforts.
% However, we claim that there are considerable unlabeled segments are underutilized in the reward function training phase.
% Therefore, how to use unlabeled segments to help obtain as much human's priori information as possible from a small number of preference-labeled segment pairs is the focus of this paper.

% But one phenomenon called ``sudden change'' in RL domain may weaken the effectiveness of CR.
% \subsection{The affect of ``sudden change'' phenemenon}
% 存在这么一个现象：在某些情况下，动作或状态的细微变化会导致整条轨迹的好坏程度发生巨大的变化。
% 在这里提前定义“轨迹”的距离(对应位置的s_t, a_t的l_2距离之和，轨迹长度不同用0补位)，好方便下面用。
% 举的例子中应该是两条轨迹很像，但结果大不相同；还有第三条轨迹，比其中一条好，比其中一条差。
% This phenemenon will result in two sequences. The effectiveness of CR will be affected. Besides, 第二个问题。
% \subsection{\emph{Similarity Trap} in PbRL}
\begin{figure}[t] %htbp
    % \vspace{-10pt}
    \centering
    \includegraphics[width=0.78\textwidth]{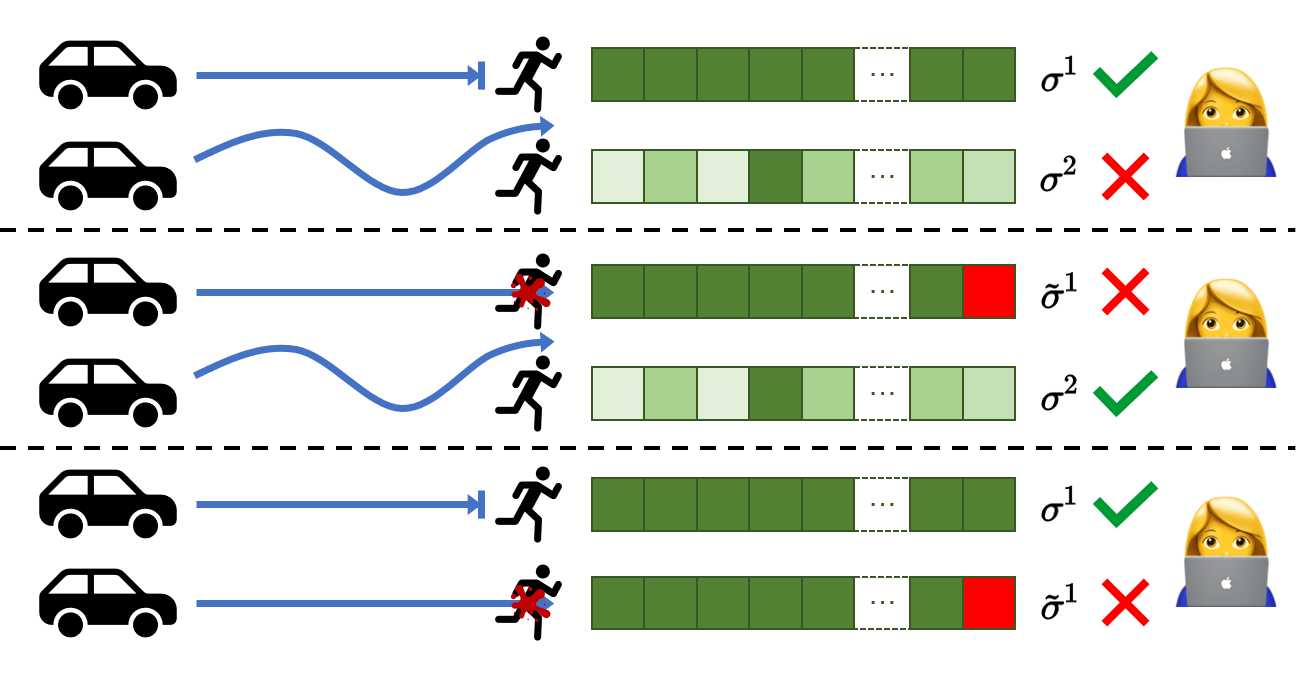}
    \caption{
        Example of \emph{Similarity Trap}.
        Two segments $\sigma^{1}$ and $\widetilde{\sigma}^{1}$ act similarly and perform equally well in most behavior (\eg driving straightly), yet the latter has a step with fatal error (\eg not avoiding pedestrians), $\sigma^{2}$ acts mediocre but conservative (\eg driving crookedly but avoiding pedestrians).
        Although there are only minor differences between $\sigma^{1}$ and $\widetilde{\sigma}^{1}$, they have diametrically opposite preferredness.
        The blocks represent the rewards that humans may confer at each step, with shades of green associated with values and red representing extremely negative rewards.
    }
    \label{fig:fatal_error}
    % \vspace{-15pt}
\end{figure}
% To improve the model generalization ability, many recent semi-supervised learning methods rely on building consistency regularization terms with unlabeled data to ensure that the model's output remains unchanged with randomly augmented inputs~\cite{DavidBerthelot2019ReMixMatchSL,QizheXie2019UnsupervisedDA}.
\textbf{Similarity Trap.}
We argue that the direct utilization of consistency regularization in PbRL is inappropriate, although it is already one of default components in most SSL methods.
In consistency regularization, the performance is substantially effected by the quality of noise addition (\ie generating neighboring pairs), but it is not always clear whether those data-augmentation methods (\eg random crop) in computer vision are suitable for segments, especially in ``feature-based'' RL\footnote{Here we slightly abuse the notation of ``feature-based'' RL to emphasize the difference from visual-based RL (states are images).}.
Consistency regularization generally assumes that the noise-added samples have the same labels as the previous ones.
But as we will discuss below, labels in PbRL are highly susceptible to small perturbations in samples.
Since the label in PbRL comes from human's preferences between two segments, it mainly conveys the information of relationship, while the label in computer vision is only related to the feature of a single input.
% So we concern that the direct utilization of self-training with consistency regularization in PbRL is inappropriate.

Based on the fact that human can be highly ``picky'' on small perturbations when providing preferences for segment pairs.
That is, the occurrence of particular action will result in one-vote veto by human for the entire segment.
As an extreme example, in the case of autonomous driving, driving straightly should be preferred to driving crookedly in most cases.
However, if the former doesn't avoid pedestrians, then it is deadly and not supposed to be preferred by humans ever, regardless of how good the previous driving behavior was (see Figure~\ref{fig:fatal_error}).
It is easy for humans to give the judgment that $\sigma^{1}\succ\sigma^{2}$ and $\widetilde{\sigma}^{1}\prec\sigma^{2}$.
In other words, similar segments can have a large gap to each other when evaluated based on different task metrics (people want comfortable driving behavior but safety is more important).
When we prepare data for PbRL, such two similar segment pairs, $x_1 = (\sigma^{1}, \sigma^{2})$ and $\widetilde{x}_1 =(\widetilde{\sigma}^{1}, \sigma^{2})$, have diametrically opposite labels $(1,0)$ and $(0,1)$, despite their similarity.
\textbf{Such data with similar samples but different labels can significantly exacerbate the difficulty of self-training with consistency regularization.
We name such phenomenon with PbRL-specific challenging data as the \emph{similarity trap}.}
Next, we elaborate the reasons for such difficulty.

\textbf{Explanation.}
For self-training in computer vision, \cite{ColinWei2021TheoreticalAO} has proved that by assuming \emph{expansion} and \emph{separation}, the fitted model will denoise the pseudolabels and achieve high accuracy on the true labels.
\emph{Expansion} assumption intuitively states that the data distribution has good continuity within each class.
On the other hand, \cite{ColinWei2021TheoreticalAO} states \emph{separation} assumption as follows (please refer to original paper for more details):
\begin{assumption}
    % \vspace{-5pt}
    (\emph{Separation}).
    \cite{ColinWei2021TheoreticalAO} assumes that $P$ is $\mathcal{B}$-separated with probability $1-\mu$ by ground-truth classifier $G^*$: $R_\mathcal{B}(G^*)\leq\mu$.
	$P$ denotes a distribution of unlabeled examples over input space $\mathcal{X}$, $\mathcal{T}$ denotes a set of transformations obtained via data augmentation, and $\mathcal{B}(x)\triangleq\left\{x^{\prime}:\exists T \in \mathcal{T} \text{such that} \left\|x^{\prime}-T(x)\right\| \leq r\right\}$ is defined as a set of points with distance $r$ from some data augmentation of $x$.
    % \vspace{-5pt}
\end{assumption}
\noindent As above, \emph{Separation} assumption states that there exist a few neighboring pairs from different classes, with a small or negligible probability $\mu$ (\eg inverse polynomial in dimension).
However, in PbRL, the presence of \emph{similarity trap} precludes the \emph{separation} assumption.
As the segment pair and its augmentation ($x_1$ and $\widetilde{x}_1$ as in the above example) are neighboring to each other but most likely from different classes.
Since the dimension of segment is extremely small compared to the image, any small perturbations have significant probability change its tendency to be preferred.
This means \emph{separation} assumption does not holds anymore in PbRL and this will weaken the effectiveness of using consistency regularization.

\textbf{Discussion.}
At the same time, we emphasize that the similarity between $\sigma^{1}$ and $\widetilde{\sigma}^{1}$ is also highly confusing, which makes the learning of $(\sigma^{1},\widetilde{\sigma}^{1})$ difficult.
Compared to the significantly different segment pairs, it is difficult to solicit information from such similar segment pairs and learn a reward function that can accurately locate step with fatal error.
Semi-supervised framework makes such difference more significant.
When the teacher model is not converged in training, it would tend to be more conservative and label $(\sigma^{1},\widetilde{\sigma}^{1})$ as $y^k=(0.5,0.5)$, while neglecting the fatal error.
When using such pseudo-label to train the student model iteratively, this can lead to further difficulties for the student to learn the desired discrimination.
As concluded in \cite{zhu2021rich}, semi-supervised learning follows the Matthew effect: ``the rich get richer''.
Consequently, \emph{similarity trap} in PbRL would cause similar segment pairs be easy to become the poor, and ``get poorer'' when conducting semi-supervised learning.

To summarize, we claim that \emph{similarity trap} in PbRL leads to two issues.
First, data augmentation on segment pairs can produce disjoint data, which makes the consistency regularization less effective.
Second, the learning of similar segment pairs is difficult although such pairs may not be proportionally significant, where this difficulty is exacerbated in semi-supervised scenarios.

\section{STRAPPER}
In this section, we present a new PbRL framework: \textbf{S}elf-\textbf{TR}aining \textbf{A}ugmented \textbf{P}reference-based learning via \textbf{PE}er \textbf{R}egularization (STRAPPER), which makes effective use of unlabeled segments by using a self-training approach and proposes peer regularization to deal with the issues induced by the \emph{similarity trap}.

% Formally, we consider a policy $\pi_\phi$, Q-function $Q_\theta$ and reward function $\widehat r_\psi$, which are updated by the following processes (see Algorithm ? for the full procedure):
% \begin{itemize}
%     \item \textbf{Step 0 (unsupervised pre-training)}: We pre-train the policy $\pi_\phi$ only using intrinsic motivation to explore and collect diverse experiences (see Section 4.1).
%     \item \textbf{Step 1 (reward learning)}: We learn a reward function $\widehat r_\psi$ that can lead to the desired behavior by getting feedback from a teacher (see Section 4.2).
%     \item \textbf{Step 2 (agent learning)}: We update the policy $\pi_\phi$ and Q-function $Q_\theta$ using an off-policy RL algorithm with relabeling to mitigate the effects of a non-stationary reward function $\widehat r_\psi$ (see Section 4.3).
%     \item Repeat \textbf{Step 1} and \textbf{Step 2}.
% \end{itemize}
\subsection{Self-training Augmented PbRL}
In STRAPPER, we first train the parameters of the teacher model $\psi^k$ on the labeled dataset $D_L$ to minimize the Eq.\ref{loss}, which we abbreviate as
\begin{equation}
    \begin{split}
        \mathcal{L}^{\text {Reward}}(\psi^k)
        =\underset{(\sigma^{i}, \sigma^{j}, y) \sim D_L}{\mathbb{E}}
        \ell(P_{\psi^k}(\sigma^{i}, \sigma^{j}),y).
    \end{split}
\end{equation}
Then, we use the teacher model to generate pseudo preferences $y^k$ for unlabeled segments pairs $(\sigma^{i}, \sigma^{j})$ which are sampled independently from the buffer $\mathcal{B}$, $y^k=\left(P_{\psi^k}[\sigma^{i} \succ \sigma^{j}],P_{\psi^k}[\sigma^{j} \succ \sigma^{i}]\right), (\sigma^{i}, \sigma^{j}) \sim D_U$.
We augment the segment $\sigma$ to $\tilde{\sigma}$ for consistency regularization, so $y^k=\left(P_{\psi^k}[\tilde{\sigma}^{i} \succ \tilde{\sigma}^{j}],P_{\psi^k}[\tilde{\sigma}^{j} \succ \tilde{\sigma}^{i}]\right), (\sigma^{i}, \sigma^{j}) \sim D_U$.
Then, the triple $(\sigma^{i}, \sigma^{j}, y^k)$ is recorded in the pseudo-labeled dataset $D_U^k$.
% Consistency regularization requires us to add noise to the process of labeling pseudo labels.
After that, the student model is trained with $D_L$ and $D_U^k$:
% As described in \cite{ColinWei2021TheoreticalAO}, if the teacher model $P_{\psi^t}$ has a classification error, training the student model $P_{\psi^{t+1}}$ directly on pseudo-labeled data set $D_U$ will not be able to correct the error, and would further deviate from the correct model.
\begin{equation}
    \begin{split}
        \mathcal{L}^{\text {Reward}}_{\mathbf{CR}}(\psi^{k+1})
        =&\underset{(\sigma^{i}, \sigma^{j}, y) \sim D_L}{\mathbb{E}}
        \ell(P_{\psi^{k+1}}(\sigma^{i}, \sigma^{j}),y) \\
        +&\underset{(\sigma^{i}, \sigma^{j}, y^k) \sim D_U^k}{\mathbb{E}}
        \ell(P_{\psi^{k+1}}(\sigma^{i}, \sigma^{j}),y^k).
    \end{split}
\end{equation}
% \begin{equation}
    %     \begin{split}
%         \mathcal{L}^{\text {Reward}}_{\mathbf{CR}}(\psi^{k+1})
%         =&\underset{(\sigma^{i}, \sigma^{j}, y) \sim D_L}{\mathbb{E}}
%         \ell(P_{\psi^{k+1}}(\sigma^{i}, \sigma^{j}),y)\\
%         +&\underset{(\sigma^{i}, \sigma^{j}, y^k) \sim D_U^k}{\mathbb{E}}
%         \ell(P_{\psi^{k+1}}(\sigma^{i}, \sigma^{j}),y^k).
%     \end{split}
% \end{equation}
For simplicity, we define the mixed dataset $D_L\cup D_U^k$ as $\tilde{D} = {(\sigma^{i}, \sigma^{j}, \tilde{y})}$, where $\tilde{y} = y,\forall{(\sigma^{i}, \sigma^{j}) \sim D_L}$, and $\tilde{y} = y^k, \forall{(\sigma^{i}, \sigma^{j}) \sim D_U}$, to simplify the above expression as:
\begin{equation}\label{CRloss}
    \begin{split}
        \mathcal{L}^{\text {Reward}}_{\mathbf{CR}}(\psi^{k+1})
        =\underset{(\sigma^{i}, \sigma^{j}, \tilde{y}) \sim \tilde{D}}{\mathbb{E}}
        \ell(P_{\psi^{k+1}}(\sigma^{i}, \sigma^{j}),\tilde{y}).
    \end{split}
\end{equation}

\subsection{Peer Regularization}
\textbf{Overall Loss Function.}
For semi-supervised problems in computer vision, it has been both empirically and theoretically proved that consistency regularization performs well.
But as discussed in Section \ref{problem}, the inherent phenomenon of \emph{similarity trap} in PbRL poses difficulties for the direct use of consistency regularization.
Due to the presence of the \emph{similarity trap}, consistency regularization improperly increases the consistency of model predictions between disjoint segment pairs and reduces the confidence in reward learning, since the augmented distribution does not match with the original one.
Inspired by \cite{YangLiu2020PeerLF}, we propose peer regularization to deal with such issue.
By adding peer regularization to Eq.\ref{CRloss}, we obtain the final loss function for training the student model as:
\begin{equation}\label{peerloss}
    \begin{split}
        \mathcal{L}^{\text {Reward}}_{\text{peer}}(\psi^{k+1})
        % =\underset{(\sigma^{i}, \sigma^{j}, \tilde{y}) \sim \tilde{D}}{\mathbb{E}}
        % \ell(P_{\psi^{k+1}}(\sigma^{i}, \sigma^{j}),\tilde{y})
        =&\ \mathcal{L}^{\text {Reward}}_{\mathbf{CR}}(\psi^{k+1}) \\
        -&\underset{\substack{{\color{red}(\sigma^{i}_{n_1},\sigma^{j}_{n_1},\tilde{y}_{n_1})} \sim \tilde{D}\\
		{\color{blue}(\sigma^{i}_{n_2},\sigma^{j}_{n_2},\tilde{y}_{n_2})} \sim \tilde{D}}}{\mathbb{E}}
        \ell\left(P_{\psi^{k+1}}{\color{red}(\sigma^{i}_{n_1}, \sigma^{j}_{n_1})}, {\color{blue}\tilde{y}_{n_2}} \right),
    \end{split}
\end{equation}
% \begin{equation}\label{peerloss}
%     \begin{split}
%         \mathcal{L}^{\text {Reward}}_{\text{peer}}(\psi&^{k+1})
%         =\underset{(\sigma^{i}, \sigma^{j}, \tilde{y}) \sim \tilde{D}}{\mathbb{E}}
%         \ell(P_{\psi^{k+1}}(\sigma^{i}, \sigma^{j}),\tilde{y})\\
%         -&\underset{\substack{(\sigma^{i}_{n_1},\sigma^{j}_{n_1},\tilde{y}_{n_1}) \sim \tilde{D}\\(\sigma^{i}_{n_2},\sigma^{j}_{n_2},\tilde{y}_{n_2}) \sim \tilde{D}}}{\mathbb{E}}
%         \ell\left(P_{\psi^{k+1}}(\sigma^{i}_{n_1}, \sigma^{j}_{n_1}),\tilde{y}_{n_2}\right),
%     \end{split}
% \end{equation}
where $(\sigma^{i}_{n_1}, \sigma^{j}_{n_1}, \tilde{y}_{n_1})$ and $(\sigma^{i}_{n_2}, \sigma^{j}_{n_2}, \tilde{y}_{n_2})$ are two independently sampled segment pairs and corresponding labels.
We call these two samples as peer samples.
The second term in Eq.\ref{peerloss}, called peer regularization, encourages the ``inconsistency'' of model predictions between random sample paires.
Intuitively, since $\tilde{y}_{n_2}$ would not provide any valuable information to label $(\sigma^{i}_{n_1}, \sigma^{j}_{n_1})$, we penalize the reward model memorizing uninformative labels.
Formally, the peer regularization leads the training to generate more confident predictions:
\begin{theorem}
    \cite{HaoCheng2021LearningWI}
    When minimizing $\mathcal{L}^{\text {Reward}}_{\text{peer}}$, the optimal solution cannot simultaneously satisfy $P_{\psi^{k+1}}[\sigma^{i} \succ \sigma^{j}] > 0$ and $ P_{\psi^{k+1}}[\sigma^{j} \succ \sigma^{i}] > 0$.
\end{theorem}
The above theorem implies that peer regularization leads to either $P_{\psi^{k+1}}[\sigma^{i} \succ \sigma^{j}]\rightarrow 1$ or $P_{\psi^{k+1}}[\sigma^{j} \succ \sigma^{i}]\rightarrow 1$, which indicates a confident prediction.
% Since reward learning in PbRL is a binary classification problem, the above theorem implies that peer regularization leads to either $P_{\psi^{k+1}}[\sigma^{i} \succ \sigma^{j}]\rightarrow 1$ or $P_{\psi^{k+1}}[\sigma^{j} \succ \sigma^{i}]\rightarrow 1$, indicating confident predictions.
This counteracts the negative impact of consistency regularization.
At the same time, confident prediction can help distinguish confusing pairs of similar segments $(\sigma^{1},\widetilde{\sigma}^{1})$, tending to pseudo-label $(0,1)$ or $(1,0)$ rather than $(0.5,0.5)$.
This would lead the pseudo-label to be closer to the real label.

\begin{algorithm}[t]
    \caption{STRAPPER}
    \label{alg:STRAPPER}
    \begin{algorithmic}[1]
        \Require number of queries $M$ and pseudo-label $N$ per feedback session
        \State Initialize parameters of $\widehat r_\psi, \pi_\phi$, a dataset for query $D_L\leftarrow\emptyset$, a pseudo-labeled dataset $D_U^k\leftarrow\emptyset$ and a replay buffer $\mathcal{B}\leftarrow\emptyset$
        % \State \emph{outerloop}:
        \State // \textsc{Reward learning}
        \For {$m$ in $1 \dots M$}
        \State Sample $\sigma^i,\sigma^j$ from $\mathcal{B}$ and query annotator for $y$, $D_L \leftarrow D_L\cup\{(\sigma^i,\sigma^j,y)\}$
        % \EndFor
        % \State \emph{innerloop}:
        % \For {each gradient step}
        \State Optimize $\mathcal{L}^{Reward}_{\text{peer}}$ in Eq.\eqref{peerloss} w.r.t $\psi$ using $\{(\sigma^i,\sigma^j,y)\}\sim D_L \cup D_U^k$
        \EndFor
        \For {$n$ in $1 \dots N$}
        \State Sample $\sigma^i,\sigma^j$ from $\mathcal{B}$ and infer the pseudo-label $y^k$ using $P_{\psi^k}$, $D_U^k \leftarrow D_U^k\cup\{(\sigma^i,\sigma^j,y^k)\}$
        \EndFor
        % \State \textbf{Goto} line 9
        % \State \textbf{Goto} \emph{innerloop}
        \State // \textsc{Policy learning}
        \For {each timestep $t$}
        \State Collect $\bs_{t+1}$ by taking $\ba_t \sim \pi_\phi(\ba_t \mid \bs_t)$, $\mathcal{B} \leftarrow \mathcal{B}\cup\{(\bs_t,\ba_t,\bs_{t+1},\widehat r_\psi(\bs_t,\ba))\}$
        \State Update $\pi_\phi$
        \EndFor
        % \State \textbf{Goto} line 3
        % \State \textbf{Goto} \emph{outerloop}
    \end{algorithmic}
\end{algorithm}
\vspace{-10pt}

\textbf{Learning with Noisy Label.}
In addition, we further describe our proposed peer regularization in terms of learning with noisy labels.
First, the \emph{similarity trap} results in a noisy pseudo-labeling process.
% So the mixed dataset in Eq.\ref{CRloss} can be considered as a noise-labeled dataset.
Because consistency regularization improperly increases the consistency of model predictions between disjoint segment pairs.
Here, the noise is introduced by the training process and therefore independent of the data distribution.
Second, PbRL is a binary classification problem.
Such two problem settings are just consistent with that in \cite{YangLiu2020PeerLF}.
Therefore, we can also straightforward use the method proposed in \cite{YangLiu2020PeerLF} for solving the problem of training with noisy labels to  derive Eq.\ref{peerloss} built on peer loss function.
% From this view, it is also straightforward to derive Eq.\ref{peerloss} built on peer loss function, \ie method proposed in \cite{YangLiu2020PeerLF} for solving the problem of training with noisy labels.

\subsection{Algorithm}
To sum up, we provide the full procedure of STRAPPER in Algorithm \ref{alg:STRAPPER}, which is given along with PEBBLE \cite{KiminLee2021PEBBLEFI}, an off-policy PbRL algorithm.

\section{Experiments}
In this section, we conduct our experiments to demonstrate the effectiveness of our proposed STRAPPER.
To verify this, we start from investigating the first question:
1) How do various SSL alternatives combined with SOTA PbRL algorithm perform? Can consistency regularization improve them?
Experiments show that current methods with consistency regularization can perform well but not on evey task due to the existence of \emph{similarity trap}.
So we secondly answer next question:
2) How does our proposed STRAPPER perform when peer regularization is further introduced to solve the problem brought by \emph{similarity trap}?
In addition, we further study the last question:
3) How sensitive is our proposed STRAPPER to the (intrinsic) noise of human preferences from non-experts?

\subsection{Setups and Details}\label{app:details}

\textbf{Simulate human annotators.}
Similar to prior work \cite{PaulFChristiano2017DeepRL, KiminLee2021PEBBLEFI}, we obtain feedback from simulated human instead of real humans.
Following \cite{KiminLee2021BPrefBP}, we first build simulated annotator to provide (rational and deterministic) preferences $y$ for queries $(\sigma^i, \sigma^j)$, \emph{oracle behavior} for short:
$$
	y = \left\{
	\begin{aligned}
		(1, 0) & \quad \text{If \ \ } \textstyle \sum_{t=1}^{H}r(\bs_t^i, \ba_t^i) > \textstyle \sum_{t=1}^{H}r(\bs_t^j, \ba_t^j) \\
		(0, 1) & \quad \text{otherwise},
	\end{aligned}
	\right.
$$
where $r(\cdot,\cdot)$ denotes the oracle reward function.

% \subsection{Experimental Details}

\begin{table}[ht]
	% \scriptsize
	% \footnotesize
	\centering
	\caption{Hyperparameters}
	\label{hyper}
	\begin{tabular}{ll}
	\hline
	\textbf{Hyperparameter}        & \textbf{Value}                     \\ \hline
	Initial temperature            & 0.1                                \\
	Length of segment              & 50                                 \\
	Learning rate                  & 0.0003 (Meta-world)                \\
								   & 0.0005 (Walker)                    \\
								   & 0.0001 (Quadruped)                 \\
	Critic target update freq      & 2                                  \\
	($\beta_1$,$\beta_2$)          & (0.9,0.999)                        \\
	Frequency of feedback          & 5000 (Meta-world)                  \\
								   & 20000 (Walker)                     \\
								   & 30000 (Quadruped)                  \\
	\# of ensemble models $N_{en}$ & 3                                  \\
	Hidden units per each layer    & 1024 (DMControl), 256 (Meta-world) \\
	\# of layers                   & 2 (DMControl), 3 (Meta-world)      \\
	Batch Size                     & 1024 (DMControl), 512 (Meta-world) \\
	Optimizer                      & Adam                               \\
	Critic EMA $\tau$              & 0.005                              \\
	Discount $\bar\gamma$          & 0.99                               \\
	Maximum budget/                & 1000/100,100/10 (DMControl)        \\
	\# of queries per session      & 10000/50,4000/20 (Meta-world)      \\
								   & 2000/25,400/10 (Meta-world)        \\
	\# of pre-training steps       & 10000                              \\ \hline
	\end{tabular}
\end{table}

\textbf{Hyper-parameters.}
The implementation of STRAPPER is based on SURF and PEBBLE, and thus inherents the hyperparameter setting from PEBBLE \cite{KiminLee2021PEBBLEFI}, which is shown in Table \ref{hyper}.
An ensemble of three reward models are initialised, while the network structures differ among environments.
The reward model for DMC consists of 2 fully-connected linear layers with 1024 neurons in each layer, whereas the one for Meta-World uses three hidden layers of 256 neurons instead.
After 10,000 pre-training steps, the reward model are trained via ADAM optimiser.
Afterwards, the model gets updated at a fixed frequency as long as the feedback budget is available.
We set the update frequency to be 5000 steps for tasks in Meta-world, while the frequencies are 20,000 steps and 30,000 steps for Walker Walk and Quadruped Walk, respectively.

For the first-time learning immeadiately after the pre-training phase, we use uniform sampling to acquire segment pairs with a segment length of 50.
To improve the feedback-efficiency for the training procedure, we switch to the disagreement-based sampling scheme for all subsequent feedback sessions.
The aim of the disagreement-based method is to find queries with high uncertainty based on the outputs among ensemble of reward models.
In each session, the number of labeled queries provided to the reward model is dependent on the training task together with the total feedback budget, and the details are specified in Table \ref{hyper}.
On the other hand, as a large portion of unlablled queries are dropped due to the confidence threshold, we sample unlabeled queries as 10 times of labeled queries. We further increase this figure to 100 if the maximum feedback budget is equal or greater than 1000.

% \textbf{Environment details.}
% For the comparison of our method to prior iterative offline RL methods~(Table~1), we consider the v0 versions of the datasets in D4RL. We take the baseline results of BEAR, BCQ, CQL, and BRAC-p from the D4RL paper~\citep{d4rl2020Fu}, and take the results of TD3+BC from their origin paper~\citep{fujimoto2021minimalist}.
% For the comparison of our method to prior non-iterative offline RL method~(Figure~3), we use the v2 versions of the dataset in D4RL. All the baseline results of behavior cloning~(BC), Decision Transform~(DT), RvS-R, and Onestep are taken from~\citet{emmons2021rvs}.

% \textbf{Evaluation protocol. }
% To evaluate models, we evaluate our results over 5 seeds (Table~1 and Figure~3).
% For each seed, instead of taking the final checkpoint produced by a training loop, we take the last $M$ ($M$ = 6 in our experiments) checkpoint models, and evaluate them over 10 episodes for each checkpoint.
% That is to say, we report the average of the evaluation scores over $5_{\text{seed}} \times 6_{\text{checkpoint}} \times 10_{\text{episode}} $ rollouts.

\textbf{Computational resources.}
Experiments are conducted by using a computational cluster with 22x GeForce RTX 2080 Ti, and 4x NVIDIA Tesla V100 32GB for 7 days.

%As discussed by PEBBLE, human annotators can provide more novel behaviors and avoid exploring a misspecified reward function. We further demonstrate the efficiency of XXX with actual human annotators (the authors).

\textbf{Benchmark tasks.}
For simulated human annotators, we take two locomotion tasks from DeepMind Control Suit \cite{tassa2018deepmind},  Walker-walk and Quadruped-walk, and four robotic manipulation tasks from Meta-world \cite{yu2021metaworld}, Window Open, Button Press, Drawer Open and Door Open.

\subsection{Different Semi-Supervised Learning Alternatives}\label{CR}

\begin{figure*}[b]
	\centering
	\includegraphics[width=0.6\textwidth]{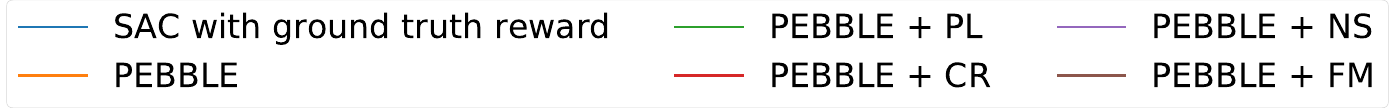} \\
	% \vspace{-5pt}
	\subfloat[Walker]{
        \includegraphics[width=0.45\textwidth]{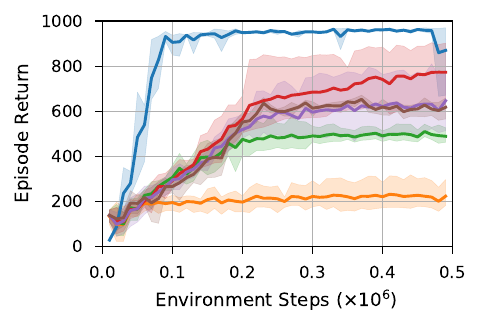}\label{fig:CR_walker}}
    \subfloat[Quadruped]{
        \includegraphics[width=0.45\textwidth]{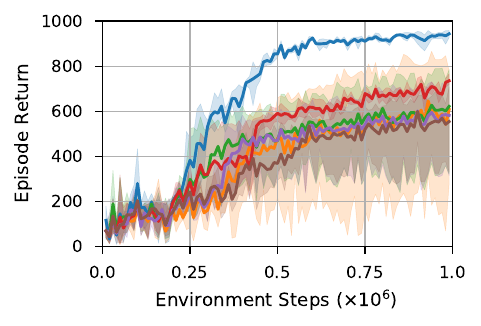}\label{fig:CR_quad}} \\
	% \vspace{-10pt}
	\subfloat[Window Open]{
		\includegraphics[width=0.45\textwidth]{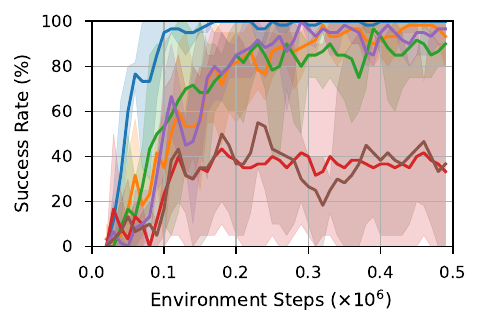}\label{fig:CR_window}}
	\subfloat[Button Press]{
		\includegraphics[width=0.45\textwidth]{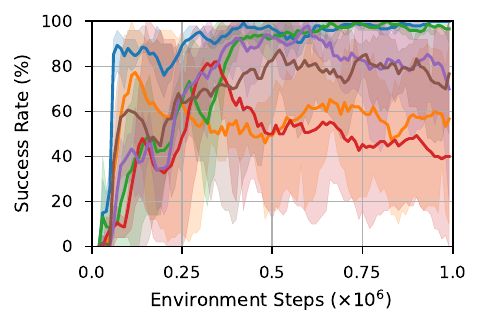}\label{fig:CR_button}}
	% \vspace{-10pt}
	\caption{
		Learning curves on four tasks as measured on the ground truth episode return for locomotion tasks and the success rate for manipulation tasks.
		The solid line and shaded regions represent the mean and standard deviation, respectively, across three runs.
		The number of feedback used in four tasks are 100, 1000, 400 and 2000.
	}
	\label{fig:CR}
	% \vspace{-10pt}
\end{figure*}

We first combine the SOTA PbRL algorithm PEBBLE \cite{KiminLee2021PEBBLEFI} with the following four alternative SSL methods:
1) \textbf{Pseudo-labeling (PL)} \cite{lee2013pseudo}:
the teacher model generates confident pseudo-labels using a fixed threshold and then updates student model using cross-entropy loss.
This is the SSL method employed in SURF \cite{park2022surf}.
2) \textbf{vanilla Consistency Regularization (CR)} \cite{xie2019unsupervised}:
constrains the output of student model between two augmented inputs using MSEloss without using pseudo-labels.
3) \textbf{FixMatch(FM)} \cite{sohn2020fixmatch}:
generates pseudo-labels with weak augmented inputs and then updates the student model with strong augmented input.
4) \textbf{self-training with Noisy Student (NS)} \cite{QizheXie2020SelfTrainingWN}:
only adds noise to the learning process of the student model.
% The temporal data augmentation methods described in \cite{park2022surf} are employed in all four methods above, while the latter three methods additionally use random amplitude scaling to augment the input.
The latter three methods all employ consistency regularization as a component and use random amplitude scaling to augment the input.
Random amplitude scaling multiplies a uniform random variable $z$ to the state, \ie $\hat{s} = s \cdot z$, where $z \sim \text{Unif}[0.995,1.005]$.
\textbf{PEBBLE} and \textbf{SAC} using the ground truth reward are also added for benchmark comparison.
% 4) Lable Smoothing (LS) follows \cite{ChristianSzegedy2016RethinkingTI}:
% we have another baseline that uses label smoothing operations on the pseudo-label of PL.

Figure \ref{fig:CR} shows SAC, PEBBLE and PEBBLE with four alternative SSL methods on locomotion tasks and robotic manipulation tasks.
As shown in Fig.\ref{fig:CR_walker} and Fig.\ref{fig:CR_quad}, all four SSL methods improve the performance of PEBBLE (orange).
And the baseline methods that utilize consistency regularization, \textbf{vanilla Consistency Regularization} (red), \textbf{self-training with Noisy Student} (purple) and \textbf{FixMatch} (brown), outperform or are competitive with \textbf{Pseudo Labeling} (green).
But when facing with robotic manipulation tasks as shown in Fig.\ref{fig:CR_window} and Fig.\ref{fig:CR_button}, these three methods perform poorly.
We will show in next subsection that STRAPPER we proposed can alleviate such problem on this type of tasks.

\subsection{STRAPPER Performs on Benchmark Experiments}
\begin{figure*}[h]
	\centering
	\vspace{-10pt}
	\includegraphics[width=0.7\textwidth]{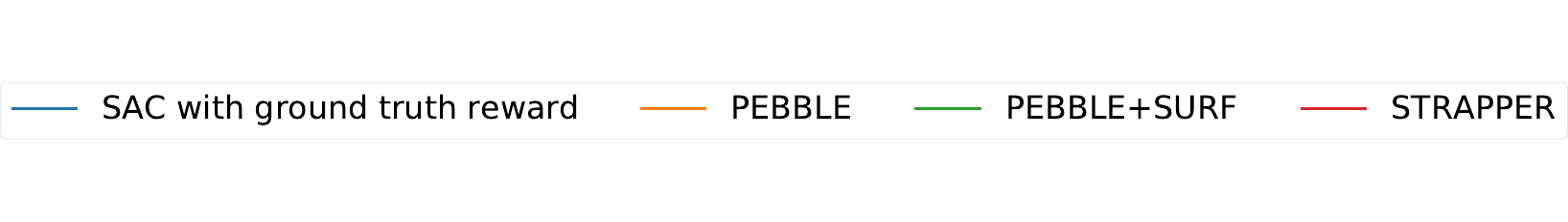} \\
	\vspace{-20pt}
	% \subfloat[Walker]{
    %     \includegraphics[width=0.48\textwidth]{exp_fig/baseline/walker_walk.pdf}}
    % % \hspace{15pt}
    % \subfloat[Quadruped]{
    %     \includegraphics[width=0.48\textwidth]{exp_fig/baseline/quad_walk.pdf}}
	\subfloat[Window Open]{
		\includegraphics[width=0.43\textwidth]{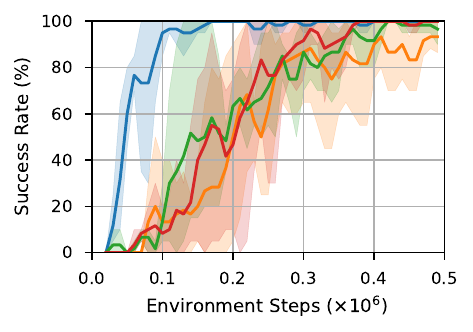}}
	\subfloat[Button Press]{
		\includegraphics[width=0.45\textwidth]{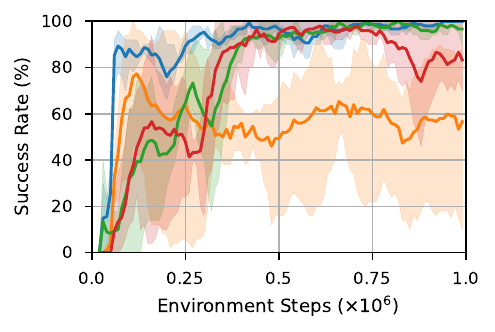}} \\
	\subfloat[Drawer Open]{\label{fig:drawer}
		\includegraphics[width=0.43\textwidth]{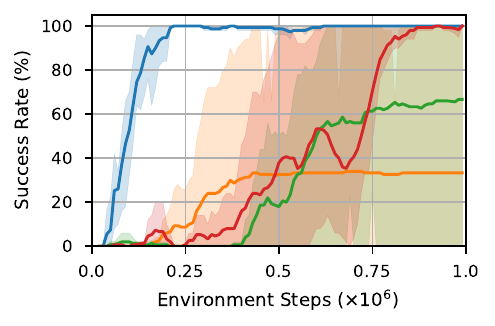}}
	\subfloat[Door Open]{\label{fig:door}
		\includegraphics[width=0.45\textwidth]{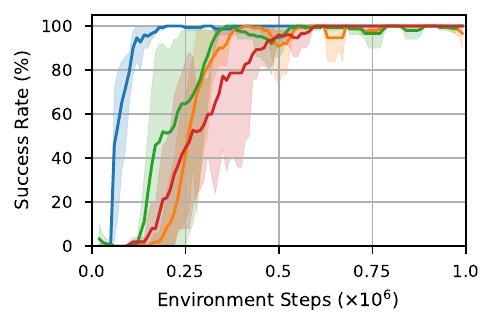}}
	% \vspace{-10pt}
	\caption{
		Learning curves on robotic manipulation tasks as measured on the ground truth success rate.
		The solid line and shaded regions represent the mean and standard deviation, respectively, across three runs.
	}
	\label{fig:benchmark}
	% \vspace{-10pt}
\end{figure*}

\textbf{Benchmark Experiments on Robotic Manipulation Tasks.}
As in Section \ref{CR}, SSL methods using consistency regularization perform well on locomotion tasks (like Walker and Quadruped) but fail on robotic manipulation tasks (like Window Open and Button Press).
We attribute this to the fact that \emph{similarity trap} occurs more frequently in robotic manipulation tasks, affecting the direct use of consistency regularization.
% in \emph{similarity trap} occurs in robotic manipulation tasks and affects the direct use of consistency regularization.
Specifically, compared to locomotion tasks, most of robotic manipulation tasks require an agent to interact with objects in the environment.
In such scenarios, certain states require precise operation, and subtle differences can bring about serious mistakes.
% As a result, \emph{similarity trap} occurs frequently and causes significant performance degradation.
Therefore we mainly focus on comparation between STRAPPER and SOTA baselines on robotic manipulation tasks.
\textbf{PEBBLE+SURF} stands for the original implementation of SURF \cite{park2022surf}, which uses \textbf{PL}, as mentioned in Section \ref{CR}.
The temporal data augmentation methods described in \cite{park2022surf} are employed in both \textbf{PEBBLE+SURF} and \textbf{STRAPPER}.
% Also it uses temporal cropping, which is also retained by us in STRAPPER.
As shown in Fig.\ref{fig:benchmark}, our method (red) outperforms or is competitive with baselines in robotic manipulation tasks.
This empirically demonstrates that the peer regularization we proposed in STRAPPER can counteract the side effect of \emph{similarity trap} when consistency regularization is used.
% Fig.\ref{fig:drawer} and Fig.\ref{fig:door} show the learning curves of STRAPPER on Drawer Open with 4,000 queries and Door Open with 4,000 queries, respectively.

\begin{figure*}[ht]
	\centering
	\includegraphics[width=0.7\textwidth]{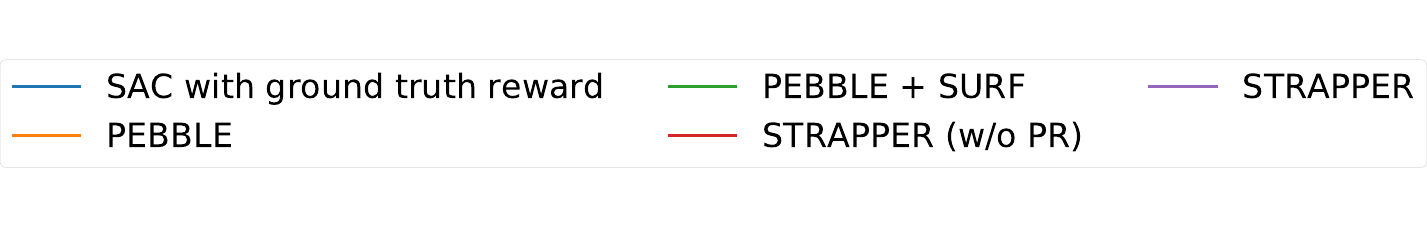} \\
	\vspace{-15pt}
	\subfloat[Walker]{
		\includegraphics[width=0.43\textwidth]{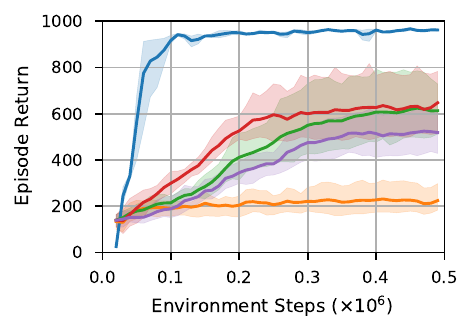}}
	\subfloat[Quadruped]{
		\includegraphics[width=0.45\textwidth]{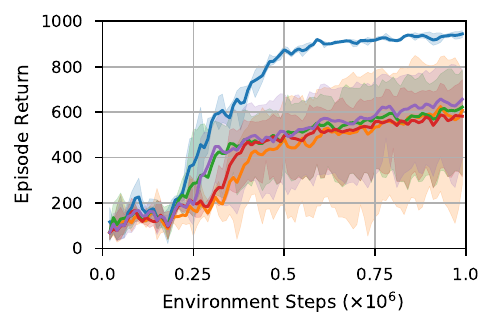}}
% 	\vspace{-10pt}
	\caption{
		Learning curves on locomotion tasks as measured on the ground truth episode return.
		The solid line and shaded regions represent the mean and standard deviation, respectively, across three runs.
	}
% 	\vspace{-10pt}
	\label{fig:strapper_dmc}
\end{figure*}

\textbf{More Benchmark Experiments on Locomotion Tasks.}
As we mention above, the baseline methods that utilizing consistency regularization without the Peer Regularization can improve the performance upon the pseudo labeling.
However, when we use Peer Regularization on such locomotion tasks, it made the performance drop, shown as \textbf{STRAPPER} in Fig.\ref{fig:strapper_dmc}.
We believe this is due to the fact that, for locomotion tasks, small differences are not enough to cause fatal errors, so the impact of \emph{similarity trap} is small.
On the contrary, the addition of Peer Regularization causes a certain degree of exploratory deficiency, which in turn leads to performance degradation.

\subsection{Experiments on Non-expert Annotators}
\begin{figure*}[t]
	\centering
	\subfloat[]{
		\begin{minipage}[b]{0.56\textwidth}
			\includegraphics[width=0.48\textwidth]{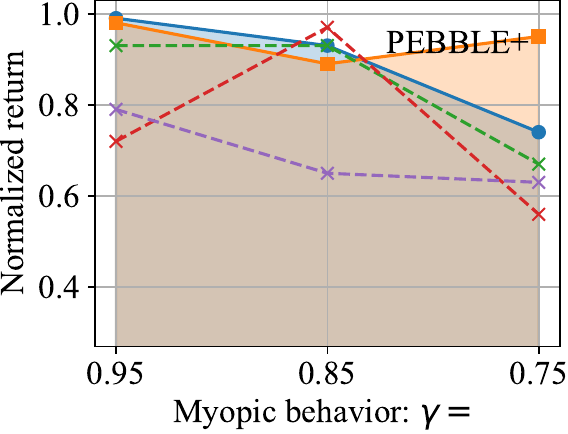}
			\includegraphics[width=0.48\textwidth]{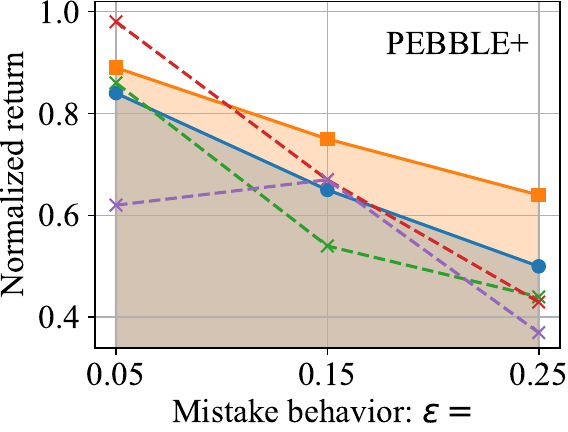} \\
			\includegraphics[width=\textwidth]{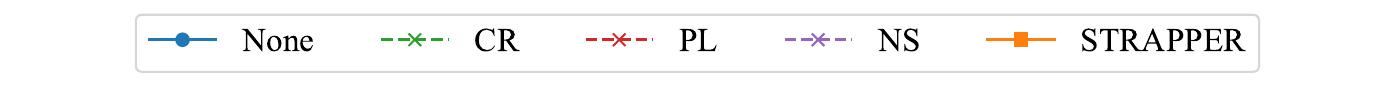}
		\end{minipage}
		\label{fig:exp:noise_plus}
		}
% 	\vspace{-10pt}
% 	\caption{
	% 	}
	\subfloat[]{
		\includegraphics[width=0.34\textwidth]{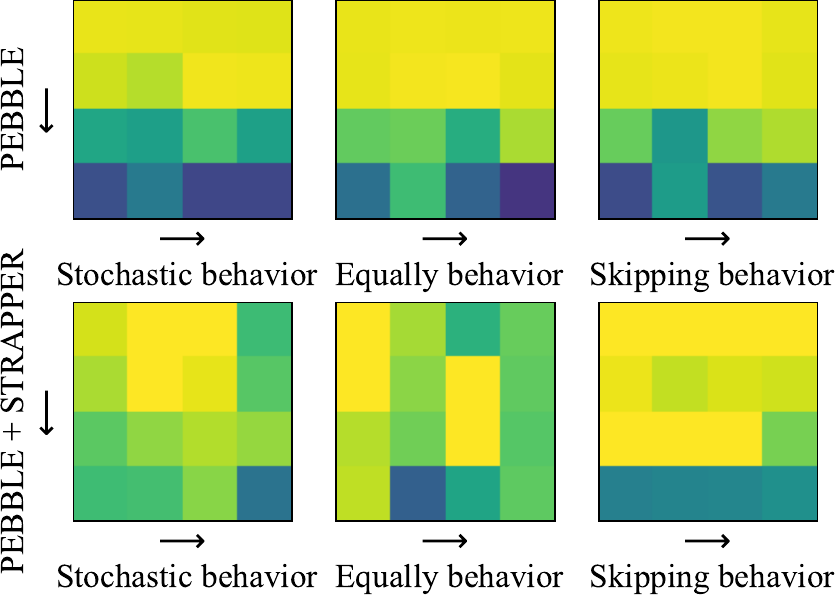}
		\includegraphics[width=0.055\textwidth]{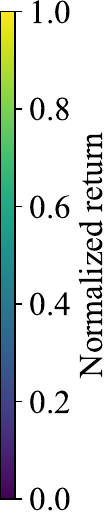}
		\label{fig:exp:noise_both}
		}
	%\vspace{-7pt}
	\caption{
		(a) Final performance of PEBBLE, and our extensions (PEBBLE + STRAPPER) on Walker-walk tasks.
		With oracle behavior (\ie rational human preferences), our STRAPPER enables more effective PbRL algorithms.
		(b) Performance under varying quantity and quality preferences in Walker-walk tasks. The down-arrow ($\downarrow$) and right-arrow ($\rightarrow$) denote that we gradually reduce the quantity and increase the noise strength with respect to the preferences, respectively.
	}
	% \vspace{-10pt}
\end{figure*}
% However, this perfectly rational and deterministic annotator is not necessarily realistic and surely does not hold for humans that constantly provide noisy preferences for some queries.
We have shown in the experiments above that the introduction of Peer Regularization in some types of environments can improve the effectiveness of consistency regularization.
We will further show that STRAPPER has additional benefits when a more realistic experimental setting is considered.
In reality, human annotators are not always perfectly rational, which may provide some stochastic preferences for query.
To address this, we generate noisy preferences using a stochastic model as:
$$
	P[\sigma^i \succ \sigma^j]
	% P[\sigma^i \succ \sigma^j; \beta, \gamma]
	= \frac{\exp(\beta\textstyle\sum_{t=1}^{H}\gamma^{H-t}r(\bs_t^i, \ba_t^i))}
	{\sum_{m\in\{i,j\}}\exp(\beta\textstyle\sum_{t=1}^{H}\gamma^{H-t}r(\bs_t^m, \ba_t^m))},
	% {\underset{m\in\{i,j\}}{\sum}\exp(\beta\textstyle\sum_{t=1}^{H}\gamma^{H-t}r(\bs_t^m, \ba_t^m))},
$$
where $\gamma \in (0,1]$ is a discount factor, modeling \emph{myopic behavior}, and $\beta \in (0, \infty]$ , modeling \emph{stochastic behavior} (expert queries become perfectly rational and deterministic as $\beta \to \infty$).
To imitate the accidental error of human expert, we flip the preference with probability of $\epsilon$, denoted as \emph{mistake behavior}.
If both segments (for query) do not contain a desired behavior, human would like to discard the query; we model this noisy behavior as \emph{skipping behavior}.
Further, if both segments have similar returns, human would like to provide a preference with $(0.5, 0.5)$; we model this behavior as \emph{equally behavior}\footnote{For more behavioral details, we refer the readers to B-Pref benchmark \cite{KiminLee2021BPrefBP}.}.
To investigate the influence of noisy preference on PbRL algorithms and study whether STRAPPER is robust to the noisy-labels, we conduct experiments with progressively enhanced noisy labels.

In Figure~\ref{fig:exp:noise_plus}, we plot the normalized return versus the strength of noise.
We observe that the straightforward PEBBLE tends to be brittle and sensitive to the noisy preferences, especially under high-level noise.
Similarly, we find that the training across three semi-supervised baselines (CR, PL, NS) can still be unstable despite of incorporating the unlabeled segments. %The performance of the same baseline varies widely across tasks.
In contrast, STRAPPER yields more robust and stable performance, which consistently outperforms the straightforward PbRL as well as semi-supervised baselines. %, except for quadruped-walk with mistake behavior ($\epsilon=0.25$).
This result serves as an evidence that unlabeled segments can be useful to eliminate the intrinsic noise in preferences.

%We perform an ablation study on the effect of unlabeled segments for our STRAPPER, resulting baseline Peer\footnote{In fact, such baseline is the straightforward combination of preference-based RL algorithms and peer loss.}. Given the same number of human preferences, Peer, compared with STRAPPER, clearly leads to worse performance; however, compared with PrefPPO and PEBBLE, Peer enables better performance. Such results, as a side-product, show that the peer loss term (\ie dis-encouraging the model memorize the uninformative labels) can help to promote the reward learning with the noisy preferences.

Further, we consider both quantity-reduced and noisy preferences in Walker-walk task.
In Figure~\ref{fig:exp:noise_both}, we illustrate the performance of PEBBLE (top row) and PEBBLE+STRAPPER (bottom row).
For PEBBLE, in this specific task, the quantity of preference seems to matter more than the quality.
By reducing the amount of preference, the performance drops significantly.
We observe that, by incorporating the unlabeled segments, STRAPPER eliminates the effect of the lack of labeled segments and thus leads to better performance, even under a high noise ratio.

\section{Discussion and Future Work}\label{future}
In this work, we show how to improve performance in PbRL using semi-supervised alternatives.
We observe a crucial phenomenon defined as \emph{similarity trap} that hinders direct applying consistency regularization to PbRL, and then propose a novel peer regularization when training the student model to fix the issue.
Empirically, we demonstrate that our proposed STRAPPER consistently outperforms prior PbRL baselines on complex locomotion and robotic manipulation tasks from DeepMind Control Suite and Meta-world.
Further, we verify that our STRAPPER is useful to eliminate the intrinsic noise in preferences.

STRAPPER has a number of limitations.
First, the current feedback acquisition still uses scripted teachers, i.e., the preference is obtained from the comparison of returns.
However, whether this is consistent with the way humans give, the preference requires further study.
Second, although semi-supervised methods improve the effectiveness of human feedback, feedback data during training is still indispensable.
This limits the large-scale application of PbRL to a certain extent.
It would be an interesting future work if the idea of Offline RL could be borrowed to train PbRL from data without feedbacks.
% First, the offline data decomposition dominates the following model-based optimization, and thus choosing suitable decomposition rule is a crucial requirement for policy inference (see experimental analysis in appendix).
% An exciting direction for future work is to study generalized task decomposition rules, including sub-trajectory augmentation, multi-goal sub-tasks detaching, and interconnected task decomposition and policy inference.
% Second, we find that when the number of sub-tasks is too large, inference is unstable, adjacent checkpoint models exhibiting larger variance on performance (such instability also exists in prior offline RL methods, discovered by Fujimoto and Gu [18]).
% One natural approach to this instability is conducting online fine-tuning (see appendix for our empirical studies).
% Going forward, we believe our work suggests a feasible alternative for generalizable offline robotic learning, by decomposing a single robotic dataset into multiple subsets, offline policy inference can benefit from performing model-based optimization.

\backmatter

\bmhead{Supplementary information}
Our code is based on repository of the PEBBLE algorithm (\burl{https://github.com/rll-research/BPref}).
We provide our source code in the supplementary material.

\begin{appendices}

% \section{Section title of first appendix}\label{secA1}

% An appendix contains supplementary information that is not an essential part of the text itself but which may be helpful in providing a more comprehensive understanding of the research problem or it is information that is too cumbersome to be included in the body of the paper.




\end{appendices}


\bibliography{ref}% common bib file


\begin{thebibliography}{45}
\ifx \bisbn   \undefined \def \bisbn  #1{ISBN #1}\fi
\ifx \binits  \undefined \def \binits#1{#1}\fi
\ifx \bauthor  \undefined \def \bauthor#1{#1}\fi
\ifx \batitle  \undefined \def \batitle#1{#1}\fi
\ifx \bjtitle  \undefined \def \bjtitle#1{#1}\fi
\ifx \bvolume  \undefined \def \bvolume#1{\textbf{#1}}\fi
\ifx \byear  \undefined \def \byear#1{#1}\fi
\ifx \bissue  \undefined \def \bissue#1{#1}\fi
\ifx \bfpage  \undefined \def \bfpage#1{#1}\fi
\ifx \blpage  \undefined \def \blpage #1{#1}\fi
\ifx \burl  \undefined \def \burl#1{\textsf{#1}}\fi
\ifx \doiurl  \undefined \def \doiurl#1{\url{https://doi.org/#1}}\fi
\ifx \betal  \undefined \def \betal{\textit{et al.}}\fi
\ifx \binstitute  \undefined \def \binstitute#1{#1}\fi
\ifx \binstitutionaled  \undefined \def \binstitutionaled#1{#1}\fi
\ifx \bctitle  \undefined \def \bctitle#1{#1}\fi
\ifx \beditor  \undefined \def \beditor#1{#1}\fi
\ifx \bpublisher  \undefined \def \bpublisher#1{#1}\fi
\ifx \bbtitle  \undefined \def \bbtitle#1{#1}\fi
\ifx \bedition  \undefined \def \bedition#1{#1}\fi
\ifx \bseriesno  \undefined \def \bseriesno#1{#1}\fi
\ifx \blocation  \undefined \def \blocation#1{#1}\fi
\ifx \bsertitle  \undefined \def \bsertitle#1{#1}\fi
\ifx \bsnm \undefined \def \bsnm#1{#1}\fi
\ifx \bsuffix \undefined \def \bsuffix#1{#1}\fi
\ifx \bparticle \undefined \def \bparticle#1{#1}\fi
\ifx \barticle \undefined \def \barticle#1{#1}\fi
\bibcommenthead
\ifx \bconfdate \undefined \def \bconfdate #1{#1}\fi
\ifx \botherref \undefined \def \botherref #1{#1}\fi
\ifx \url \undefined \def \url#1{\textsf{#1}}\fi
\ifx \bchapter \undefined \def \bchapter#1{#1}\fi
\ifx \bbook \undefined \def \bbook#1{#1}\fi
\ifx \bcomment \undefined \def \bcomment#1{#1}\fi
\ifx \oauthor \undefined \def \oauthor#1{#1}\fi
\ifx \citeauthoryear \undefined \def \citeauthoryear#1{#1}\fi
\ifx \endbibitem  \undefined \def \endbibitem {}\fi
\ifx \bconflocation  \undefined \def \bconflocation#1{#1}\fi
\ifx \arxivurl  \undefined \def \arxivurl#1{\textsf{#1}}\fi
\csname PreBibitemsHook\endcsname

%%% 1
\bibitem{NateKohl2004PolicyGR}
\begin{bchapter}
\bauthor{\bsnm{Kohl}, \binits{N.}},
\bauthor{\bsnm{Stone}, \binits{P.}}:
\bctitle{Policy gradient reinforcement learning for fast quadrupedal
  locomotion}.
In: \bbtitle{IEEE International Conference on Robotics and Automation, 2004.
  Proceedings. ICRA '04. 2004}
(\byear{2004}).
\burl{https://doi.org/10.1109/robot.2004.1307456}
\end{bchapter}
\endbibitem

%%% 2
\bibitem{JensKober2008PolicySF}
\begin{bchapter}
\bauthor{\bsnm{Kober}, \binits{J.}},
\bauthor{\bsnm{Peters}, \binits{J.}}:
\bctitle{Policy search for motor primitives in robotics}.
In: \bbtitle{Neural Information Processing Systems}
(\byear{2008})
\end{bchapter}
\endbibitem

%%% 3
\bibitem{JensKober2013ReinforcementLI}
\begin{barticle}
\bauthor{\bsnm{Kober}, \binits{J.}},
\bauthor{\bsnm{Bagnell}, \binits{J.A.}},
\bauthor{\bsnm{Peters}, \binits{J.}}:
\batitle{Reinforcement learning in robotics: {A} survey}.
\bjtitle{The International Journal of Robotics Research}
\bvolume{32}(\bissue{11}),
\bfpage{1238}--\blpage{1274}
(\byear{2013}).
\doiurl{10.1177/0278364913495721}
\end{barticle}
\endbibitem

%%% 4
\bibitem{DavidSilver2017MasteringTG}
\begin{barticle}
\bauthor{\bsnm{Silver}, \binits{D.}},
\bauthor{\bsnm{Schrittwieser}, \binits{J.}},
\bauthor{\bsnm{Simonyan}, \binits{K.}},
\bauthor{\bsnm{Antonoglou}, \binits{I.}},
\bauthor{\bsnm{Huang}, \binits{A.}},
\bauthor{\bsnm{Guez}, \binits{A.}},
\bauthor{\bsnm{Hubert}, \binits{T.}},
\bauthor{\bsnm{Baker}, \binits{L.}},
\bauthor{\bsnm{Lai}, \binits{M.}},
\bauthor{\bsnm{Bolton}, \binits{A.}},
\bauthor{\bsnm{Chen}, \binits{Y.}},
\bauthor{\bsnm{Lillicrap}, \binits{T.}},
\bauthor{\bsnm{Hui}, \binits{F.}},
\bauthor{\bsnm{Sifre}, \binits{L.}},
\bauthor{\bparticle{van~den} \bsnm{Driessche}, \binits{G.}},
\bauthor{\bsnm{Graepel}, \binits{T.}},
\bauthor{\bsnm{Hassabis}, \binits{D.}}:
\batitle{Mastering the game of go without human knowledge}.
\bjtitle{Nature}
\bvolume{550}(\bissue{7676}),
\bfpage{354}--\blpage{359}
(\byear{2017}).
\doiurl{10.1038/nature24270}
\end{barticle}
\endbibitem

%%% 5
\bibitem{DmitryKalashnikov2018QTOptSD}
\begin{bchapter}
\bauthor{\bsnm{Kalashnikov}, \binits{D.}},
\bauthor{\bsnm{Irpan}, \binits{A.}},
\bauthor{\bsnm{Pastor}, \binits{P.}},
\bauthor{\bsnm{Ibarz}, \binits{J.}},
\bauthor{\bsnm{Herzog}, \binits{A.}},
\bauthor{\bsnm{Jang}, \binits{E.}},
\bauthor{\bsnm{Quillen}, \binits{D.}},
\bauthor{\bsnm{Holly}, \binits{E.}},
\bauthor{\bsnm{Kalakrishnan}, \binits{M.}},
\bauthor{\bsnm{Vanhoucke}, \binits{V.}},
\bauthor{\bsnm{Levine}, \binits{S.}}:
\bctitle{{QT}-opt: {Scalable} deep reinforcement learning for vision-based
  robotic manipulation}.
In: \bbtitle{Computer Vision and Pattern Recognition}
(\byear{2018})
\end{bchapter}
\endbibitem

%%% 6
\bibitem{OriolVinyals2019GrandmasterLI}
\begin{barticle}
\bauthor{\bsnm{Vinyals}, \binits{O.}},
\bauthor{\bsnm{Babuschkin}, \binits{I.}},
\bauthor{\bsnm{Czarnecki}, \binits{W.M.}},
\bauthor{\bsnm{Mathieu}, \binits{M.}},
\bauthor{\bsnm{Dudzik}, \binits{A.}},
\bauthor{\bsnm{Chung}, \binits{J.}},
\bauthor{\bsnm{Choi}, \binits{D.H.}},
\bauthor{\bsnm{Powell}, \binits{R.}},
\bauthor{\bsnm{Ewalds}, \binits{T.}},
\bauthor{\bsnm{Georgiev}, \binits{P.}},
\bauthor{\bsnm{Oh}, \binits{J.}},
\bauthor{\bsnm{Horgan}, \binits{D.}},
\bauthor{\bsnm{Kroiss}, \binits{M.}},
\bauthor{\bsnm{Danihelka}, \binits{I.}},
\bauthor{\bsnm{Huang}, \binits{A.}},
\bauthor{\bsnm{Sifre}, \binits{L.}},
\bauthor{\bsnm{Cai}, \binits{T.}},
\bauthor{\bsnm{Agapiou}, \binits{J.P.}},
\bauthor{\bsnm{Jaderberg}, \binits{M.}},
\bauthor{\bsnm{Vezhnevets}, \binits{A.S.}},
\bauthor{\bsnm{Leblond}, \binits{R.}},
\bauthor{\bsnm{Pohlen}, \binits{T.}},
\bauthor{\bsnm{Dalibard}, \binits{V.}},
\bauthor{\bsnm{Budden}, \binits{D.}},
\bauthor{\bsnm{Sulsky}, \binits{Y.}},
\bauthor{\bsnm{Molloy}, \binits{J.}},
\bauthor{\bsnm{Paine}, \binits{T.L.}},
\bauthor{\bsnm{Gulcehre}, \binits{C.}},
\bauthor{\bsnm{Wang}, \binits{Z.}},
\bauthor{\bsnm{Pfaff}, \binits{T.}},
\bauthor{\bsnm{Wu}, \binits{Y.}},
\bauthor{\bsnm{Ring}, \binits{R.}},
\bauthor{\bsnm{Yogatama}, \binits{D.}},
\bauthor{\bsnm{Wünsch}, \binits{D.}},
\bauthor{\bsnm{McKinney}, \binits{K.}},
\bauthor{\bsnm{Smith}, \binits{O.}},
\bauthor{\bsnm{Schaul}, \binits{T.}},
\bauthor{\bsnm{Lillicrap}, \binits{T.}},
\bauthor{\bsnm{Kavukcuoglu}, \binits{K.}},
\bauthor{\bsnm{Hassabis}, \binits{D.}},
\bauthor{\bsnm{Apps}, \binits{C.}},
\bauthor{\bsnm{Silver}, \binits{D.}}:
\batitle{Grandmaster level in {StarCraft} {II} using multi-agent reinforcement
  learning}.
\bjtitle{Nature}
\bvolume{575}(\bissue{7782}),
\bfpage{350}--\blpage{354}
(\byear{2019}).
\doiurl{10.1038/s41586-019-1724-z}
\end{barticle}
\endbibitem

%%% 7
\bibitem{DarioAmodei2016ConcretePI}
\begin{botherref}
\oauthor{\bsnm{Amodei}, \binits{D.}},
\oauthor{\bsnm{Olah}, \binits{C.}},
\oauthor{\bsnm{Steinhardt}, \binits{J.}},
\oauthor{\bsnm{Christiano}, \binits{P.F.}},
\oauthor{\bsnm{Schulman}, \binits{J.}},
\oauthor{\bsnm{Man{\'e}}, \binits{D.}}:
Concrete problems in {AI} safety.
arXiv: Artificial Intelligence
(2016)
\end{botherref}
\endbibitem

%%% 8
\bibitem{RohinShah2019PreferencesII}
\begin{botherref}
\oauthor{\bsnm{Shah}, \binits{R.}},
\oauthor{\bsnm{Krasheninnikov}, \binits{D.}},
\oauthor{\bsnm{Alexander}, \binits{J.}},
\oauthor{\bsnm{Abbeel}, \binits{P.}},
\oauthor{\bsnm{Dragan}, \binits{A.D.}}:
Preferences implicit in the state of the world.
arXiv: Learning
(2019)
\end{botherref}
\endbibitem

%%% 9
\bibitem{AlexanderMattTurner2020AvoidingSE}
\begin{botherref}
\oauthor{\bsnm{Turner}, \binits{A.M.}},
\oauthor{\bsnm{Ratzlaff}, \binits{N.}},
\oauthor{\bsnm{Tadepalli}, \binits{P.}}:
Avoiding side effects in complex environments.
arXiv: Artificial Intelligence
(2020)
\end{botherref}
\endbibitem

%%% 10
\bibitem{PaulFChristiano2017DeepRL}
\begin{botherref}
\oauthor{\bsnm{Christiano}, \binits{P.F.}},
\oauthor{\bsnm{Leike}, \binits{J.}},
\oauthor{\bsnm{Brown}, \binits{T.B.}},
\oauthor{\bsnm{Martic}, \binits{M.}},
\oauthor{\bsnm{Legg}, \binits{S.}},
\oauthor{\bsnm{Amodei}, \binits{D.}}:
Deep reinforcement learning from human preferences.
arXiv: Machine Learning
(2017)
\end{botherref}
\endbibitem

%%% 11
\bibitem{ErdemByk2018BatchAP}
\begin{botherref}
\oauthor{\bsnm{Bıyık}, \binits{E.}},
\oauthor{\bsnm{Sadigh}, \binits{D.}}:
Batch active preference-based learning of reward functions.
arXiv: Learning
(2018)
\end{botherref}
\endbibitem

%%% 12
\bibitem{DorsaSadigh2017ActivePL}
\begin{bchapter}
\bauthor{\bsnm{Sadigh}, \binits{D.}},
\bauthor{\bsnm{Dragan}, \binits{A.}},
\bauthor{\bsnm{Sastry}, \binits{S.}},
\bauthor{\bsnm{Seshia}, \binits{S.}}:
\bctitle{Active preference-based learning of reward functions}.
In: \bbtitle{Robotics: Science and Systems XIII}
(\byear{2017}).
\burl{https://doi.org/10.15607/rss.2017.xiii.053}
\end{bchapter}
\endbibitem

%%% 13
\bibitem{ErdemByk2020ActivePG}
\begin{bchapter}
\bauthor{\bsnm{Biyik}, \binits{E.}},
\bauthor{\bsnm{Huynh}, \binits{N.}},
\bauthor{\bsnm{Kochenderfer}, \binits{M.}},
\bauthor{\bsnm{Sadigh}, \binits{D.}}:
\bctitle{Active preference-based {Gaussian} process regression for reward
  learning}.
In: \bbtitle{Robotics: Science and Systems XVI}
(\byear{2020}).
\burl{https://doi.org/10.15607/rss.2020.xvi.041}
\end{bchapter}
\endbibitem

%%% 14
\bibitem{KiminLee2021BPrefBP}
\begin{bchapter}
\bauthor{\bsnm{Lee}, \binits{K.}},
\bauthor{\bsnm{Smith}, \binits{L.}},
\bauthor{\bsnm{Dragan}, \binits{A.}},
\bauthor{\bsnm{Abbeel}, \binits{P.}}:
\bctitle{B-pref: {Benchmarking} preference-based reinforcement learning}.
(\byear{2021})
\end{bchapter}
\endbibitem

%%% 15
\bibitem{shin2021offline}
\begin{botherref}
\oauthor{\bsnm{Shin}, \binits{D.}},
\oauthor{\bsnm{Brown}, \binits{D.S.}}:
Offline preference-based apprenticeship learning.
arXiv preprint arXiv:2107.09251
(2021)
\end{botherref}
\endbibitem

%%% 16
\bibitem{KiminLee2021PEBBLEFI}
\begin{bchapter}
\bauthor{\bsnm{Lee}, \binits{K.}},
\bauthor{\bsnm{Smith}, \binits{L.M.}},
\bauthor{\bsnm{Abbeel}, \binits{P.}}:
\bctitle{{PEBBLE:} {Feedback-efficient} interactive reinforcement learning via
  relabeling experience and unsupervised pre-training}.
In: \bbtitle{International Conference on Machine Learning}
(\byear{2021})
\end{bchapter}
\endbibitem

%%% 17
\bibitem{liang2021reward}
\begin{bchapter}
\bauthor{\bsnm{Liang}, \binits{X.}},
\bauthor{\bsnm{Shu}, \binits{K.}},
\bauthor{\bsnm{Lee}, \binits{K.}},
\bauthor{\bsnm{Abbeel}, \binits{P.}}:
\bctitle{Reward uncertainty for exploration in preference-based reinforcement
  learning}.
In: \bbtitle{Deep RL Workshop NeurIPS 2021}
(\byear{2021})
\end{bchapter}
\endbibitem

%%% 18
\bibitem{BorjaIbarz2018RewardLF}
\begin{bchapter}
\bauthor{\bsnm{Ibarz}, \binits{B.}},
\bauthor{\bsnm{Leike}, \binits{J.}},
\bauthor{\bsnm{Pohlen}, \binits{T.}},
\bauthor{\bsnm{Irving}, \binits{G.}},
\bauthor{\bsnm{Legg}, \binits{S.}},
\bauthor{\bsnm{Amodei}, \binits{D.}}:
\bctitle{Reward learning from human preferences and demonstrations in {Atari}}.
In: \bbtitle{Neural Information Processing Systems}
(\byear{2018})
\end{bchapter}
\endbibitem

%%% 19
\bibitem{chapelle2009semi}
\begin{barticle}
\bauthor{\bsnm{Chapelle}, \binits{O.}},
\bauthor{\bsnm{Scholkopf}, \binits{B.}},
\bauthor{\bsnm{Zien}, \binits{A.} \bsuffix{Eds.}}:
\batitle{Semi-supervised learning {(Chapelle,} o. et al., eds.; 2006) {[Book}
  reviews]}.
\bjtitle{IEEE Trans. Neural Netw.}
\bvolume{20}(\bissue{3}),
\bfpage{542}--\blpage{542}
(\byear{2009}).
\doiurl{10.1109/tnn.2009.2015974}
\end{barticle}
\endbibitem

%%% 20
\bibitem{park2022surf}
\begin{botherref}
\oauthor{\bsnm{Park}, \binits{J.}},
\oauthor{\bsnm{Seo}, \binits{Y.}},
\oauthor{\bsnm{Shin}, \binits{J.}},
\oauthor{\bsnm{Lee}, \binits{H.}},
\oauthor{\bsnm{Abbeel}, \binits{P.}},
\oauthor{\bsnm{Lee}, \binits{K.}}:
Surf: Semi-supervised reward learning with data augmentation for
  feedback-efficient preference-based reinforcement learning.
arXiv preprint arXiv:2203.10050
(2022)
\end{botherref}
\endbibitem

%%% 21
\bibitem{laine2016temporal}
\begin{botherref}
\oauthor{\bsnm{Laine}, \binits{S.}},
\oauthor{\bsnm{Aila}, \binits{T.}}:
Temporal ensembling for semi-supervised learning.
arXiv preprint arXiv:1610.02242
(2016)
\end{botherref}
\endbibitem

%%% 22
\bibitem{sajjadi2016regularization}
\begin{botherref}
\oauthor{\bsnm{Sajjadi}, \binits{M.}},
\oauthor{\bsnm{Javanmardi}, \binits{M.}},
\oauthor{\bsnm{Tasdizen}, \binits{T.}}:
Regularization with stochastic transformations and perturbations for deep
  semi-supervised learning.
Advances in neural information processing systems
\textbf{29}
(2016)
\end{botherref}
\endbibitem

%%% 23
\bibitem{xie2019unsupervised}
\begin{botherref}
\oauthor{\bsnm{Xie}, \binits{Q.}},
\oauthor{\bsnm{Dai}, \binits{Z.}},
\oauthor{\bsnm{Hovy}, \binits{E.}},
\oauthor{\bsnm{Luong}, \binits{M.-T.}},
\oauthor{\bsnm{Le}, \binits{Q.V.}}:
Unsupervised data augmentation for consistency training.
arXiv preprint arXiv:1904.12848
(2019)
\end{botherref}
\endbibitem

%%% 24
\bibitem{YangLiu2020PeerLF}
\begin{bchapter}
\bauthor{\bsnm{Liu}, \binits{Y.}},
\bauthor{\bsnm{Guo}, \binits{H.}}:
\bctitle{Peer loss functions: {Learning} from noisy labels without knowing
  noise rates}.
In: \bbtitle{International Conference on Machine Learning}
(\byear{2020})
\end{bchapter}
\endbibitem

%%% 25
\bibitem{DilipArumugam2019DeepRL}
\begin{botherref}
\oauthor{\bsnm{Arumugam}, \binits{D.}},
\oauthor{\bsnm{Lee}, \binits{J.K.}},
\oauthor{\bsnm{Saskin}, \binits{S.}},
\oauthor{\bsnm{Littman}, \binits{M.L.}}:
Deep reinforcement learning from policy-dependent human feedback.
arXiv: Learning
(2019)
\end{botherref}
\endbibitem

%%% 26
\bibitem{WBradleyKnox2009InteractivelySA}
\begin{bchapter}
\bauthor{\bsnm{Knox}, \binits{W.B.}},
\bauthor{\bsnm{Stone}, \binits{P.}}:
\bctitle{Interactively shaping agents via human reinforcement}.
In: \bbtitle{Proceedings of the Fifth International Conference on Knowledge
  Capture - K-CAP '09}
(\byear{2009}).
\burl{https://doi.org/10.1145/1597735.1597738}
\end{bchapter}
\endbibitem

%%% 27
\bibitem{GarrettWarnell2017DeepTI}
\begin{botherref}
\oauthor{\bsnm{Warnell}, \binits{G.}},
\oauthor{\bsnm{Waytowich}, \binits{N.R.}},
\oauthor{\bsnm{Lawhern}, \binits{V.J.}},
\oauthor{\bsnm{Stone}, \binits{P.}}:
Deep {TAMER:} {Interactive} agent shaping in high-dimensional state spaces.
arXiv: Artificial Intelligence
(2017)
\end{botherref}
\endbibitem

%%% 28
\bibitem{DavidYarowsky1995UNSUPERVISEDWS}
\begin{bchapter}
\bauthor{\bsnm{Yarowsky}, \binits{D.}}:
\bctitle{Unsupervised word sense disambiguation rivaling supervised methods}.
In: \bbtitle{Proceedings of the 33rd Annual Meeting on Association for
  Computational Linguistics -}
(\byear{1995}).
\burl{https://doi.org/10.3115/981658.981684}
\end{bchapter}
\endbibitem

%%% 29
\bibitem{KamalNigam2000AnalyzingTE}
\begin{bchapter}
\bauthor{\bsnm{Nigam}, \binits{K.}},
\bauthor{\bsnm{Ghani}, \binits{R.}}:
\bctitle{Analyzing the effectiveness and applicability of co-training}.
In: \bbtitle{Proceedings of the Ninth International Conference on Information
  and Knowledge Management - CIKM '00}
(\byear{2000}).
\burl{https://doi.org/10.1145/354756.354805}
\end{bchapter}
\endbibitem

%%% 30
\bibitem{zhu2005semi}
\begin{botherref}
\oauthor{\bsnm{Zhu}, \binits{X.J.}}:
Semi-supervised learning literature survey
(2005)
\end{botherref}
\endbibitem

%%% 31
\bibitem{CharlesJRosenberg2005SemiSupervisedSO}
\begin{bchapter}
\bauthor{\bsnm{Rosenberg}, \binits{C.}},
\bauthor{\bsnm{Hebert}, \binits{M.}},
\bauthor{\bsnm{Schneiderman}, \binits{H.}}:
\bctitle{Semi-supervised self-training of object detection models}.
In: \bbtitle{2005 Seventh IEEE Workshops on Applications of Computer Vision
  (WACV/MOTION'05) - Volume 1}
(\byear{2005}).
\burl{https://doi.org/10.1109/acvmot.2005.107}
\end{bchapter}
\endbibitem

%%% 32
\bibitem{lee2013pseudo}
\begin{bchapter}
\bauthor{\bsnm{Lee}, \binits{D.-H.}}, \betal:
\bctitle{Pseudo-label: {The} simple and efficient semi-supervised learning
  method for deep neural networks}.
In: \bbtitle{Workshop on Challenges in Representation Learning, ICML},
vol. \bseriesno{3},
p. \bfpage{896}
(\byear{2013})
\end{bchapter}
\endbibitem

%%% 33
\bibitem{bachman2014learning}
\begin{barticle}
\bauthor{\bsnm{Bachman}, \binits{P.}},
\bauthor{\bsnm{Alsharif}, \binits{O.}},
\bauthor{\bsnm{Precup}, \binits{D.}}:
\batitle{Learning with pseudo-ensembles}.
\bjtitle{Adv Neural Inf Process Syst}
\bvolume{27},
\bfpage{3365}--\blpage{3373}
(\byear{2014})
\end{barticle}
\endbibitem

%%% 34
\bibitem{rasmus2015semi}
\begin{botherref}
\oauthor{\bsnm{Rasmus}, \binits{A.}},
\oauthor{\bsnm{Berglund}, \binits{M.}},
\oauthor{\bsnm{Honkala}, \binits{M.}},
\oauthor{\bsnm{Valpola}, \binits{H.}},
\oauthor{\bsnm{Raiko}, \binits{T.}}:
Semi-supervised learning with ladder networks.
Advances in neural information processing systems
\textbf{28}
(2015)
\end{botherref}
\endbibitem

%%% 35
\bibitem{sohn2020fixmatch}
\begin{botherref}
\oauthor{\bsnm{Sohn}, \binits{K.}},
\oauthor{\bsnm{Berthelot}, \binits{D.}},
\oauthor{\bsnm{Li}, \binits{C.-L.}},
\oauthor{\bsnm{Zhang}, \binits{Z.}},
\oauthor{\bsnm{Carlini}, \binits{N.}},
\oauthor{\bsnm{Cubuk}, \binits{E.D.}},
\oauthor{\bsnm{Kurakin}, \binits{A.}},
\oauthor{\bsnm{Zhang}, \binits{H.}},
\oauthor{\bsnm{Raffel}, \binits{C.}}:
{FixMatch:} {Simplifying} Semi-Supervised Learning with Consistency and
  Confidence
(2020)
\end{botherref}
\endbibitem

%%% 36
\bibitem{QizheXie2020SelfTrainingWN}
\begin{bchapter}
\bauthor{\bsnm{Xie}, \binits{Q.}},
\bauthor{\bsnm{Luong}, \binits{M.-T.}},
\bauthor{\bsnm{Hovy}, \binits{E.}},
\bauthor{\bsnm{Le}, \binits{Q.V.}}:
\bctitle{Self-training with noisy student improves {ImageNet} classification}.
In: \bbtitle{2020 IEEE/CVF Conference on Computer Vision and Pattern
  Recognition (CVPR)}
(\byear{2020}).
\burl{https://doi.org/10.1109/cvpr42600.2020.01070}
\end{bchapter}
\endbibitem

%%% 37
\bibitem{zhu2021rich}
\begin{botherref}
\oauthor{\bsnm{Zhu}, \binits{Z.}},
\oauthor{\bsnm{Luo}, \binits{T.}},
\oauthor{\bsnm{Liu}, \binits{Y.}}:
The Rich Get Richer: {Disparate} Impact of Semi-Supervised Learning
(2021)
\end{botherref}
\endbibitem

%%% 38
\bibitem{RichardSutton1988ReinforcementLA}
\begin{bchapter}
\bauthor{\bsnm{Sutton}, \binits{R.}},
\bauthor{\bsnm{Barto}, \binits{A.G.}}:
\bctitle{Reinforcement learning: {An} introduction}.
(\byear{1988})
\end{bchapter}
\endbibitem

%%% 39
\bibitem{JanLeike2018ScalableAA}
\begin{botherref}
\oauthor{\bsnm{Leike}, \binits{J.}},
\oauthor{\bsnm{Krueger}, \binits{D.}},
\oauthor{\bsnm{Everitt}, \binits{T.}},
\oauthor{\bsnm{Martic}, \binits{M.}},
\oauthor{\bsnm{Maini}, \binits{V.}},
\oauthor{\bsnm{Legg}, \binits{S.}}:
Scalable agent alignment via reward modeling: {A} research direction.
arXiv: Learning
(2018)
\end{botherref}
\endbibitem

%%% 40
\bibitem{AaronWilson2012ABA}
\begin{bchapter}
\bauthor{\bsnm{Wilson}, \binits{A.}},
\bauthor{\bsnm{Fern}, \binits{A.}},
\bauthor{\bsnm{Tadepalli}, \binits{P.}}:
\bctitle{A {Bayesian} approach for policy learning from trajectory preference
  queries}.
In: \bbtitle{Neural Information Processing Systems}
(\byear{2012})
\end{bchapter}
\endbibitem

%%% 41
\bibitem{Bradley1952RankAO}
\begin{barticle}
\bauthor{\bsnm{Bradley}, \binits{R.A.}},
\bauthor{\bsnm{Terry}, \binits{M.E.}}:
\batitle{Rank analysis of incomplete block designs: {I.} the method of paired
  comparisons}.
\bjtitle{Biometrika}
\bvolume{39}(\bissue{3/4}),
\bfpage{324}
(\byear{1952}).
\doiurl{10.2307/2334029}
\end{barticle}
\endbibitem

%%% 42
\bibitem{ColinWei2021TheoreticalAO}
\begin{bchapter}
\bauthor{\bsnm{Wei}, \binits{C.}},
\bauthor{\bsnm{Shen}, \binits{K.}},
\bauthor{\bsnm{Chen}, \binits{Y.}},
\bauthor{\bsnm{Ma}, \binits{T.}}:
\bctitle{Theoretical analysis of self-training with deep networks on unlabeled
  data}.
In: \bbtitle{International Conference on Learning Representations}
(\byear{2021})
\end{bchapter}
\endbibitem

%%% 43
\bibitem{HaoCheng2021LearningWI}
\begin{bchapter}
\bauthor{\bsnm{Zhu}, \binits{Z.}},
\bauthor{\bsnm{Liu}, \binits{T.}},
\bauthor{\bsnm{Liu}, \binits{Y.}}:
\bctitle{A second-order approach to learning with instance-dependent label
  noise}.
In: \bbtitle{2021 IEEE/CVF Conference on Computer Vision and Pattern
  Recognition (CVPR)}
(\byear{2021}).
\burl{https://doi.org/10.1109/cvpr46437.2021.00998}
\end{bchapter}
\endbibitem

%%% 44
\bibitem{tassa2018deepmind}
\begin{botherref}
\oauthor{\bsnm{Tassa}, \binits{Y.}},
\oauthor{\bsnm{Doron}, \binits{Y.}},
\oauthor{\bsnm{Muldal}, \binits{A.}},
\oauthor{\bsnm{Erez}, \binits{T.}},
\oauthor{\bsnm{Li}, \binits{Y.}},
\oauthor{\bparticle{de} \bsnm{Las~Casas}, \binits{D.}},
\oauthor{\bsnm{Budden}, \binits{D.}},
\oauthor{\bsnm{Abdolmaleki}, \binits{A.}},
\oauthor{\bsnm{Merel}, \binits{J.}},
\oauthor{\bsnm{Lefrancq}, \binits{A.}},
\oauthor{\bsnm{Lillicrap}, \binits{T.}},
\oauthor{\bsnm{Riedmiller}, \binits{M.}}:
{DeepMind} Control Suite
(2018)
\end{botherref}
\endbibitem

%%% 45
\bibitem{yu2021metaworld}
\begin{botherref}
\oauthor{\bsnm{Yu}, \binits{T.}},
\oauthor{\bsnm{Quillen}, \binits{D.}},
\oauthor{\bsnm{He}, \binits{Z.}},
\oauthor{\bsnm{Julian}, \binits{R.}},
\oauthor{\bsnm{Narayan}, \binits{A.}},
\oauthor{\bsnm{Shively}, \binits{H.}},
\oauthor{\bsnm{Bellathur}, \binits{A.}},
\oauthor{\bsnm{Hausman}, \binits{K.}},
\oauthor{\bsnm{Finn}, \binits{C.}},
\oauthor{\bsnm{Levine}, \binits{S.}}:
Meta-World: {A} Benchmark and Evaluation for Multi-Task and Meta Reinforcement
  Learning
(2021)
\end{botherref}
\endbibitem

\end{thebibliography}


\end{document}